\documentclass[journal]{IEEEtran}
\usepackage{amsmath,amsfonts}
\usepackage{array}
\usepackage{textcomp}
\usepackage{stfloats}
\usepackage{url}
\usepackage{verbatim}
\usepackage{graphicx}
\usepackage{subfigure}
\usepackage{cite}
\usepackage{bm}
\usepackage{todonotes}
\usepackage[ruled,linesnumbered,vlined]{algorithm2e}
\usepackage{multirow}
\usepackage{threeparttable}
\usepackage{hyperref}
\usepackage{xcolor}

\usepackage[normalem]{ulem}
\newcommand{\rev}[2]{#2}

\begin{document}

\title{\rev{}{Co-Optimization of Tool Orientations, Kinematic Redundancy, and Waypoint Timing for Robot-Assisted Manufacturing}}

\author{Yongxue Chen, Tianyu Zhang,~\IEEEmembership{Member,~IEEE}, Yuming Huang, Tao Liu, and Charlie C.L. Wang,~\IEEEmembership{Senior Member,~IEEE} 
}

\markboth{Authors' version under review}
{Chen \MakeLowercase{\textit{et al.}}: Concurrent and Scalable Trajectory Optimization}


\maketitle

\begin{abstract}
\rev{}{In this paper, we present a concurrent and scalable trajectory optimization method to improve the quality of robot-assisted manufacturing. Our method simultaneously optimizes tool orientations, kinematic redundancy, and waypoint timing on input toolpaths with large numbers of waypoints to improve kinematic smoothness while incorporating manufacturing constraints.} Differently, existing methods always determine them in a decoupled manner. To deal with the large number of waypoints on a toolpath, we propose a decomposition-based numerical scheme to optimize the trajectory in an out-of-core manner, which can also run in parallel to improve the efficiency. Simulations and physical experiments have been conducted to demonstrate the performance of our method in examples of robot-assisted additive manufacturing. 
\end{abstract}

\def\abstractname{Note to Practitioners}
\begin{abstract}
In robot-assisted manufacturing, how to determine the motion commands according to a sequence of waypoints is a typical problem to be solved where the waypoints represent the positions of a tool-tip. Factors in three aspects need to be planned at each waypoint, including the tool orientation, the tool speed and the redundant degrees-of-freedom on the robotic system. In the trajectory planning step, the objective is always defined as improving the kinematic performance of joint motion in terms of velocity, acceleration, and jerk. Taking the strategy of existing methods that consider these aspects separately will generate less optimal results. This paper presents a new formulation that optimizes all these together while assigning certain manufacturing constraints. Considering that a toolpath can consist of a large number of waypoints in practice, how to improve planning efficiency with limited computer memory is an important issue to be solved. A decomposition based numerical scheme is developed to tackle this problem. The aforementioned issues can be effectively solved by the method proposed in this paper, the performance of which has been demonstrated on a dual robotic system with 6+2 DoFs. The proposed method is general and can also be applied to other types of systems with single or multiple robots as well as other manufacturing methods (e.g. milling).
\end{abstract}


\section{Introduction}\label{secIntro}
\IEEEPARstart{U}{sing} robots can provide a large workspace and high flexibility in manufacturing, which therefore facilitate the fabrication of product designs with larger dimensions and more complicated geometry. For example, robot-assisted \textit{additive manufacturing} (AM) enables 3D printing of free-form curved layers, providing the benefits of support-structure free \cite{Dai2018}, reinforced mechanical strength \cite{Fang2020ACM}, and improved surface quality \cite{Etienne2019}. These multiple benefits have recently been realized on the same models \textcolor{blue}{(refs.~\cite{Zhang2022ACM,Liu2024})}.

\rev{}{Motion and trajectory planning} is crucial in robot-assisted manufacturing, as it directly affects the quality of resultant workpieces. Several issues need to be addressed. First of all, \rev{}{most manufacturing tasks require only five \textit{degrees-of-freedom} (DoF), whereas robotic systems typically have higher DoFs, leading to kinematic redundancy. It means that, for any given task-satisfying constraints of the end-effector, a robotic system can have an infinite number of \textit{inverse kinematics} (IK) solutions. Additionally, toolpaths} provided for manufacturing \rev{}{tasks} often only place strict requirements on the positions of the tooltip. That means both the tool orientations and the time-sequence can be adjusted while satisfying certain manufacturing constraints. Furthermore, in both AM and \textit{subtractive manufacturing} (SM), the kinematic smoothness is to be optimized for ensuring the quality of material processing. When determining the joint trajectory of robots \rev{}{with kinematic redundancy,} all these factors need to be considered simultaneously. On the other hand, for models with complicated geometry, the number of waypoints for a toolpath can go up to more than 8k, which is difficult to solve directly by existing methods (e.g., \cite{Peng2020SciCN,Lu2021RCIM,Chen2023JMSE}). \rev{}{We address these challenges in this paper by proposing a new method that can simultaneously optimize tool orientations, kinematic redundancy of the robotic system, and waypoint timing for toolpaths with numerous waypoints.}

\subsection{Related Work}
We review the relevant literature for tool orientations, kinematic redundancy and kinematic smoothness below.

Tool orientations in multi-axis AM are often selected according to the surface normals of curved layers for achieving good material adhesion \textcolor{blue}{\cite{Liu2024,Fang2020ACM,Zhang2022ACM,Zhang2021RAL}}. For SM such as \textit{computer numerical control} (CNC) milling, the methods to optimize tool orientations have been extensively studied \cite{Lu2017,Chiou2005,Fard2010,GONG2010,WANG2021}. A general purpose of these methods is to maximize the machining width while avoiding gouging and singularity.

After determining the tool orientation, the 5-DoF toolpaths are obtained. In robot-assisted manufacturing, a most common situation is to execute the 5-DoF toolpath by a 6-DoF robot. In this case, the rotational angle of the end-effector around the tool axis is a redundant DoF and needs to be planned. Some researchers proposed to solve the redundancy at each waypoint one by one to improve the system's local performance such as the robotic stiffness \cite{Xiong2019RCIM,Liao2022TMech,Guo2015RCIM}, contour error \cite{Lin2022RCIM}, and surface location error \cite{Hou2023JMSE}. In order to improve global metrics of the manufacturing trajectory, such as smoothness and energy efficiency, some global optimization methods \cite{Lu2021RCIM,Lyu2017} were presented to solve the redundant angles at all waypoints simultaneously. 

To enhance the kinematics of the trajectory, some methods proposed optimization criteria based on the joint velocity \cite{Shibata1997} and acceleration \cite{Peng2020SciCN}. A local filtering method was proposed in \cite{Dai2020} to minimize the jerk of a trajectory. This method is not limited to the 6-DoF robot and can also be applied to a dual-robot system in experiments. This method can only consider a small number of waypoints at each iteration, and its computational efficiency is very low when dealing with long toolpaths. 

\rev{}{Recently, methods have been proposed to optimize the tool orientation and the robot redundancy simultaneously (e.g., \cite{Chen2023JMSE,Lu2022Tmech,Liao2020RCIM,Li2022RCIM,Xu2022,MA2023RCIM,LIAO2024RCIM,Malhan2023Tase}).}
Compared with the decoupled planning strategy, these methods can find more optimal solutions. Among these methods, the optimization algorithm based on \textit{sequential quadratic programming} (SQP) \cite{Chen2023JMSE} has shown its high efficiency and robustness in dealing with non-linear problems. It has been generalized and demonstrated to plan flank milling motions for a 6-DoF robot \cite{Chen2024RCIM}. \rev{}{However, these methods cannot incorporate constraints related to waypoint timing, such as joint velocity, acceleration, and jerk, necessitating additional speed planning.}

The result of tool orientation and robot redundancy planning is a sequence of the robotic joint angles. With the determined joint angles, studies on speed planning have been conducted to improve the jerks of joint angles and the total manufacturing time (ref.~\cite{Wen2023TASE,Chen2021SCICN}). These methods can be classified into three categories, including the dynamic programming method \cite{Oberherber2015MS}, the numerical integration method \cite{Barnett2021TRO}, and the convex optimization methods \textcolor{blue}{\cite{Dong2017ASME,Hauser2014IJRR,Fan2013IJAMT,Consolini2019TRO,Chen2021SCICN,Zhao2015,Ji2024TASE,Proia2023TASE}}. 

The aforementioned methods are effective for trajectory planning in general, but they also have the following limitations.
\begin{itemize}
    \item {Most of these methods only focus on one or two of the factors that have an influence on the quality of the resultant trajectory, which means that either the problems are solved in a manner of multiple stages or by simply fixing some of these variables. The generation of real optimal results is prevented.}
    
    \item {For most of the above methods, when toolpaths consist of a large number of waypoints, the computational efficiency is low and the required computer memory is enormous.}
    
    \item {Most methods for tool orientation and redundancy planning are designed for SM with a single 6-DoF robot, and they cannot be directly extended to cover the manufacturing systems with multiple robots.}
\end{itemize}
A concurrent and scalable trajectory optimization method for \rev{}{a dual-robot system with kinematic redundancy} is needed.

\subsection{Our Method}
First of all, a kinematic model for dual robotic arms is formulated. We then introduce the kinematic smoothness metrics based on joint velocity, acceleration, and jerk. Constraints on orientations and tool-tip motion are modeled according to the requirements for successful additive manufacturing. On this basis, an optimization method is developed to concurrently plan tool orientation, \rev{}{kinematic redundancy}, and \rev{}{waypoint timing}. The objective function of optimization is defined to enhance the kinematic smoothness of the resultant trajectory. To solve this problem for toolpaths with a large number of waypoints, an efficient computing scheme is proposed based on SQP and a developed decomposition strategy.

The contributions of the paper are mainly in the following areas.
\begin{itemize}
    \item {A new formulation with tool orientation, robot redundancy, and manufacturing time-sequence being optimized concurrently so that results with better performance can be achieved.}
    
    \item {A decomposition-based numerical scheme to optimize the trajectory in an out-of-core manner which can also run in parallel to improve the efficiency.}

    \item {A general formulation supporting both the single and the dual robots systems by incorporating the manufacturing constraints for AM applications.}
\end{itemize}

The rest of our paper is organized as follows. Section \ref{Sec:ProblemStatement} presents the kinematic model of the robotic system and formulates the trajectory optimization problem in detail. The efficient method of numerical computation is presented in Sec.~\ref{Sec:TrajOpt}. Section \ref{sec:discussion} discusses details of the implementation and special cases. Simulations and experimental results are presented in Sec.~\ref{sec:validation}. Finally, we conclude the work in Sec.~\ref{sec:conclusion}.

\section{Problem Statement} \label{Sec:ProblemStatement}

\subsection{Kinematic Model of the 6+2 DoFs Dual-Robot System}
\label{sec:kineModel}

\begin{figure}[!t]
    \centering
    \subfigure[]{\includegraphics[width=.278\textwidth]{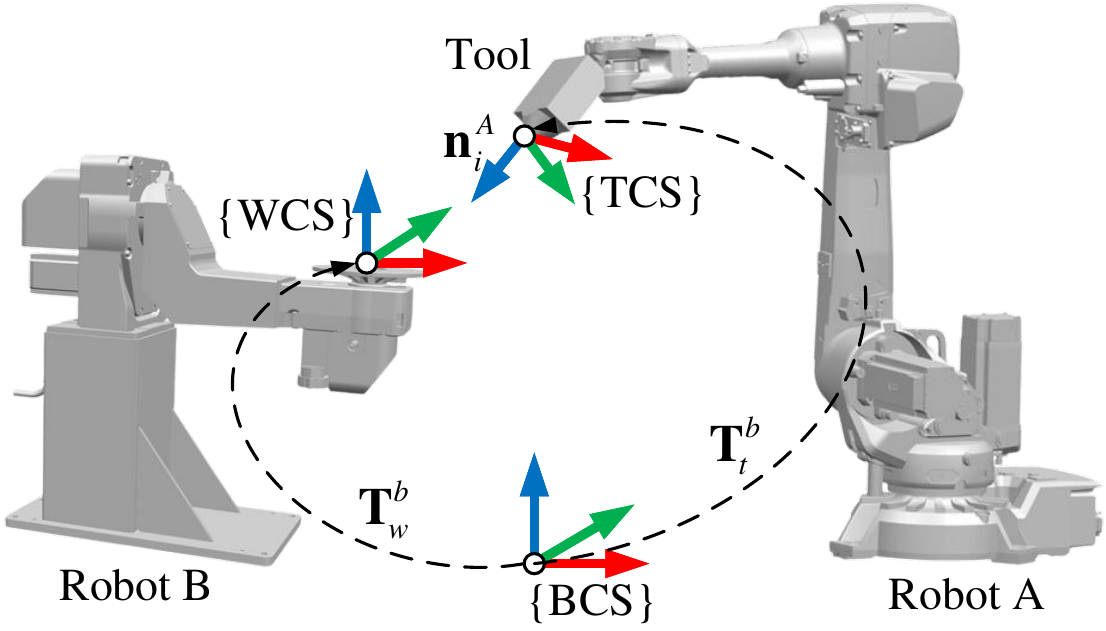}}
    \
    \subfigure[]{\includegraphics[width=.198\textwidth]{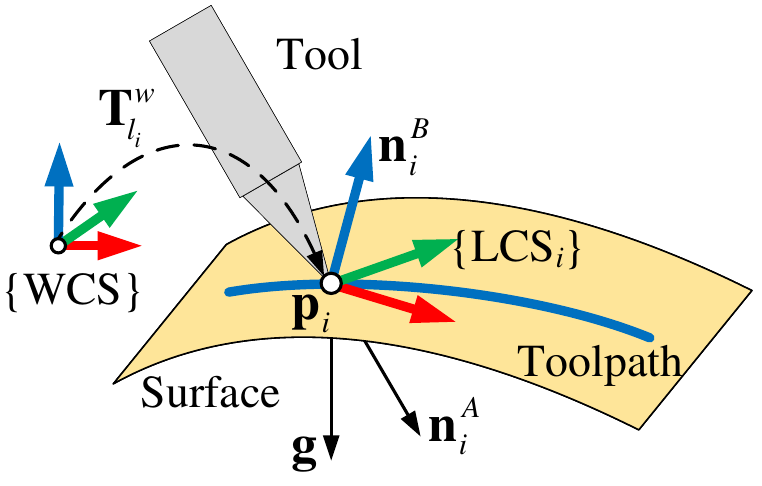}}
    \caption{The kinematic model of a robotic manufacturing system: (a) the dual-robots system with 6+2 DoFs, and (b) the schematic of the local coordinate systems around each waypoint $\mathbf{p}_i$.}
    \label{fig:km}
\end{figure}

Dual-robot systems have been increasingly used in different manufacturing scenarios \textcolor{blue}{\cite{LIAO2024RCIM,Sui2023,Fang2024,XU2024}}. In particular, dual-robot systems allow the orientation change of both the workpiece and the printer head, which makes it easier to accumulate materials from more flexible directions in the model space enabling the fabrication of complex curved layers. This section presents the kinematic model of the 6+2 DoFs dual-robot system as shown in Fig. \ref{fig:km}(a). Note that the modeling framework as well as the trajectory optimization method presented in this paper can be employed for a variety of single and dual-robot systems (i.e., not limited to this particular setup).

The toolpath for manufacturing consists of a series of discrete waypoints denoted by $\{\textbf{p}_i \, | \, i=1,2,\cdots,n\}$ with $n$ being the number of waypoints, which is usually generated by a path planner and employed as the input of our trajectory optimizer. The toolpath must be accurately followed by the tooltip. The kinematic model presented here is for the case when the tooltip is located at $\textbf{p}_i$, where the toolpath $\{\textbf{p}_i\}$ is represented in the workpiece coordinate system \{WCS\}. 

The robot holding the printer head is denoted as Robot A. The tool coordinate system \{TCS\} is established at the tooltip. Its $z$-axis is assigned along the tool axis being denoted as $\textbf{n}^A_i$. The kinematic model from the base coordinate system \{BCS\} to \{TCS\} is
\begin{equation}
    \label{eq:kmR}
    \textbf{T}^b_t ( \textbf{q}^A_i )=
    \begin{bmatrix}
        \textbf{R}^b_t(\bm{\omega}_i) & \textbf{p}^b_{t,i} \\
        \textbf{0} & 1
    \end{bmatrix},
\end{equation}
where $\textbf{T}^b_t$ represents the forward kinematics (FK) of Robot A, and $\textbf{q}^A_i \in \mathbb{R}^6$ is the vector of its joint angles for reaching the waypoint $\textbf{p}_i$. $\textbf{p}^b_{t,i} \in \mathbb{R}^3$ and $\textbf{R}^b_t \in SO(3)$ represent the tooltip position and tool orientation in \{BCS\} respectively. $\bm{\omega}_i=[\omega_{i,1},\omega_{i,2},\omega_{i,3}]^T \in \mathbb{R}^3$ is the exponential coordinate of $\textbf{R}^b_t$ (i.e., $\textbf{R}^b_t=exp([{\bm{\omega}_i}])$) with
\begin{equation*}
    [{\bm{\omega}_i}] = 
    \begin{bmatrix}
    0 & -\omega_{i,3} & \omega_{i,2}\\
    \omega_{i,3} & 0 & -\omega_{i,1}\\
    -\omega_{i,2} & \omega_{i,1} & 0
    \end{bmatrix}.
\end{equation*}
In this paper, the operator $[\cdot]$ acting on a vector in $\mathbb{R}^3$ gives a skew-symmetric matrix belonging to the lie algebra $so(3)$. For a given $\textbf{p}^b_{t,i}$, $\bm{\omega}_i$ can be considered as an independent variable in Eq.\eqref{eq:kmR}. In other words, for any $\bm{\omega}_i$, as long as the homogeneous transformation matrix on the right-hand side of Eq.\eqref{eq:kmR} is reachable, we can get $\textbf{q}^A_i$ through the IK of the robot.

As shown in Fig. \ref{fig:km}(b), \{LCS$_i$\} denotes a local coordinate system constructed at $\textbf{p}_i$ with its $z$-axis being assigned along the normal $\textbf{n}^B_i$ of the current layer. For the robot holding the workpiece, denoted as Robot B, the transformation from \{BCS\} to \{LCS$_i$\} can then be formulated as
\begin{equation}
    \label{eq:kmP}
    \textbf{T}^b_w (\textbf{q}^B_i) \textbf{T}^{w}_{l_i}
    =
    \begin{bmatrix}
        \textbf{R}^b_i & \textbf{p}^b_i \\
        \textbf{0} & 1
    \end{bmatrix},
\end{equation}
where $\textbf{T}^b_w$ is the FK of Robot B, representing the position and orientation of the workpiece coordinate system \{WCS\} with respect to (w.r.t.) \{BCS\}, and $\textbf{q}^B_i \in \mathbb{R}^2$ is the vector of joint angles for Robot B. $\textbf{T}^{w}_{l_i}$ represent the pose of \{LCS$_i$\} w.r.t. \{WCS\}. $\textbf{R}^b_i$ indicates the orientation of \{LCS$_i$\} w.r.t. \{BCS\}, and $\textbf{p}^b_i$ gives the representation of $\textbf{p}_i$ in \{BCS\}. Obviously, $\textbf{q}^B_i$ is an independent variable in Eq.\eqref{eq:kmP} since $\textbf{T}^{w}_{l_i}$ is constant.

A successful manufacturing process requires the tooltip being located at the desired waypoints, which gives
\begin{equation}
    \label{eq:kmCons}
    \textbf{p}^b_{t,i} = \textbf{p}^b_i
\end{equation}
for $i=1, 2, \cdots, n$. To ensure the quality of additive manufacturing, more constraints are also placed for tool orientations and other kinematic metrics with details given in Sec.\ref{sec:manuRequirement}.

Based on the above discussion, Eqs.\eqref{eq:kmR}-\eqref{eq:kmCons} define the kinematic model of the dual-robot system for a waypoint $\textbf{p}_i$ on the toolpath. \rev{}{Although this model can be constructed without introducing \{BCS\}, it is included here for two main reasons: 1) to incorporate manufacturing constraints depending on the direction of gravity (see Sec.~\ref{sec:manuRequirement} for more details) that is defined in \{BCS\}, and 2) to provide a generalized method for constructing kinematic models, facilitating the replacement of any robot with ones in different configurations.}

There are five independent variables in this model, which are $\bm{\theta}_i=[ \bm{\omega}_i^T,(\textbf{q}^B_i)^T ]^T \in \mathbb{R}^5$. $\{\bm{\theta}_i \, | \, i=1,2,\cdots ,n \}$ are used as the optimization variables together with the manufacturing time-sequence discussed below for the trajectory optimization problem. Details will presented in Sec.~\ref{sec:optProblem}.

\subsection{Manufacturing Time-Sequence as Variables}
\label{sec:time-sequence}
The time spent by the tool moving along the toolpath is defined as a manufacturing time-sequence $\{t_i \, | \, i=2,\cdots,n \}$, which can also be optimized to improve the quality and efficiency of manufacturing. Specifically, $t_i$ represents the time taken for the tool moving from $\mathbf{p}_{i-1}$ to $\mathbf{p}_i$. Therefore, the total manufacturing time for an input toolpath can be obtained by
\begin{equation}
    \Phi_{tm} = \sum_{i=2}^{n} t_i.
\end{equation}
When changing the time-sequence at all waypoints, the velocity of the tooltip is planned therefore also other kinematic metrics.  

\subsection{Kinematic Metric}
The kinematic smoothness of a trajectory has been observed as an important way to improve the quality of robot-assisted manufacturing. It is mainly measured by the velocity, acceleration, and jerk of all the robot joints. Given the vector of joint angles of the dual-robot system as $\textbf{q}_i = [(\textbf{q}^A_i)^T, (\textbf{q}^B_i)^T]^T$ at the waypoint $\textbf{p}_i$, the corresponding velocity, acceleration, and jerk vectors of joints are denoted by $\textbf{v}_i$, $\textbf{a}_i$, and $\textbf{j}_i$.

As the value of $t_i$ can change for different $i$, the unevenly spaced numerical differentiation is used to evaluate $\textbf{v}_i$, $\textbf{a}_i$, and $\textbf{j}_i$ (ref.~\cite{Gautschi2011}). We can have the velocity as 
\begin{equation}
    \label{eq:v_i}
    \textbf{v}_i= \frac {t_i^2 \textbf{q}_{i+1} + (t_{i+1}^2 - t_i^2) \textbf{q}_i - t_{i+1}^2 \textbf{q}_{i-1}} { (t_{i+1} + t_i) t_i t_{i+1} }. 
\end{equation}
The acceleration $\textbf{a}_i$ and the jerk $\textbf{j}_i$ can also derived by the method of \cite{Gautschi2011} -- see Appendix \ref{AppA} for the formulas.

\rev{}{We propose the following metric $\Phi_{smooth}$ for evaluating the overall kinematic smoothness of a robotic system, which is defined as the integral of the norms of velocity, acceleration, and jerk along the toolpath.}
\begin{equation}
    \label{eq:phi}
    \Phi_{smooth} = \sum_{i=1}^n \phi_i \Delta s_i
\end{equation}
\begin{equation}
    \label{eq:phi_i}
    \phi_i = k_v \textbf{v}_i^T \textbf{W} \textbf{v}_i + k_a \textbf{a}_i^T \textbf{W} \textbf{a}_i + k_j \textbf{j}_i^T \textbf{W} \textbf{j}_i
\end{equation}
where $\Delta s_i = (s_i + s_{i+1})/2$ with $s_i=\|\mathbf{p}_i - \mathbf{p}_{i-1} \|$ being the distance between $\mathbf{p}_i$ and $\mathbf{p}_{i-1}$, $k_v$, $k_a$ and $k_j$ are three non-negative coefficients, and $\mathbf{W}$ is a diagonal non-negative matrix used to control the importance of each joint. We choose the values of $k_v$, $k_a$ and $k_j$ after normalization (see Sec.~\ref{subsecNorm} for details). 
$\mathbf{W} = \mathbf{I}$ is also simply chosen for all examples tested in our work. 

\subsection{Manufacturing Requirements} \label{sec:manuRequirement}
The requirement of the relative orientation between the tool and the workpiece has been studied in \cite{Lu2017,Chiou2005,Fard2010,GONG2010,Chen2023JMSE} for subtractive manufacturing such as milling. However, due to the working principle of material extrusion in additive manufacturing, constraints on the absolute orientation of both the printer head and the workpiece w.r.t. \{BCS\} need to be imposed in addition to the relative orientation. 

\subsubsection{Requirements on Orientations}
At each waypoint $\mathbf{p}_i$, the constraints on orientations are defined by the extrusion direction $\textbf{n}^A_i$, the normal $\textbf{n}^B_i$ of curved layer, and the gravity direction $\textbf{g}$. The schematic of these three directions is shown in Fig.~\ref{fig:km}(b). 

    

\begin{figure}[!t]
\centering
\includegraphics[width=0.46 \textwidth]{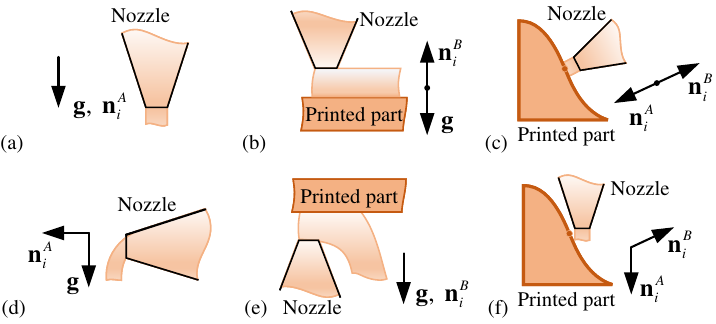}
\caption{The influence of orientations on additive manufacturing: the ideal conditions are given in (a)-(c) and the corresponding failure cases are shown in (d)-(f).}
\label{fig:ThreeDirections}
\end{figure}

Extruding materials out of the nozzle of a printer head needs the help of gravity. If the extrusion direction $\textbf{n}^A_i$ deviates too much from the gravity $\textbf{g}$, the material flow will be non-uniform and not follow the direction of $\textbf{n}^A_i$. The quality of material deposition will be influenced as shown in Fig.\ref{fig:ThreeDirections}(d). The deviation is controlled by a constraint $\textbf{n}^A_i \cdot \textbf{g} \geq \cos{\alpha}$. As listed in Eq.\eqref{cons:nNozzle}, $\textbf{n}^A_i$ is determined by $\bm{\omega}_i$ as it is the third column of $\textbf{R}^b_t$ in Eq.\eqref{eq:kmR}.

To ensure a successful stacking of the material, the angle between the layer normal $\textbf{n}^B_i$ and $\textbf{g}$ should be constrained. If this is not controlled, an extreme case is as illustrated in Fig.\ref{fig:ThreeDirections}(e) where the extruded material will not be able to bond onto the already printed part -- i.e., the material will fall away caused by gravity. The constraint is given in Eq.\eqref{cons:nNormal}, where $\beta$ is a threshold angle smaller than $\pi/2$ and $\textbf{n}^B_i$ is determined by $\textbf{q}^B_i$ as it is the third column of $\textbf{R}^b_i$ in Eq.\eqref{eq:kmP}.

At the same time, the angle between $\mathbf{n}^B_i$ and $\mathbf{n}^A_i$ also needs to be constrained. Since there is a certain distance between the tooltip and the printed layer in practice, a too-large deviation between these two directions will prevent the material from reaching the desired position as illustrated in Fig.\ref{fig:ThreeDirections}(f). This constraint is given in Eq.\eqref{cons:nNN} with a threshold angle $\gamma$. 

\subsubsection{Requirements on Tooltip Motion}
The motion at the tooltip needs also to be controlled to ensure the quality of manufacturing. For example, if the tooltip goes through a sharp turn at a breakneck speed w.r.t. \{WCS\}, there is a high possibility that the extruded material will be carried away by the printer head and cannot be stably located in the desired position. To avoid this problem, the constraints of the tooltip velocity $v^t_i$, the tangential acceleration $a^t_i$, and the normal acceleration $a^n_i$ at the tooltip are imposed in \{WCS\} to ensure effective material adhesion at all waypoints. At a waypoint $\textbf{p}_i$, the speed of motion $v^t_i$ is 
\begin{equation}
    \label{eq:vt}
    v^t_i=s_i/t_i.
\end{equation}
\rev{}{With the detailed derivation given in Appendix \ref{AppB}), the tangential acceleration $a^t_i$ and the normal acceleration $a^n_i$ can be computed by}
\begin{equation}
    \label{eq:at}
    a^{t}_i = \frac{2(s_{i+1} t_i -s_i t_{t+1})} {t_i t_{i+1} (t_i + t_{i+1})},
\end{equation}
\begin{equation}
    \label{eq:an}
    a^{n}_i = \kappa_i \frac
    {s_i^2 t_{i+1}^2 +s_{i+1}^2 t_i^2 -2 s_i s_{i+1} t_i t_{i+1} \textup{cos}(\sigma_i) } 
    {4 t_i^2 t_{i+1}^2},
\end{equation}
where $s_i = \| \mathbf{p}_i - \mathbf{p}_{i-1}\|$, $\kappa_i$ is the approximated curvature of the toolpath at the $\textbf{p}_i$, and $\sigma_i$ is the angle between the two vectors $\textbf{p}_i \textbf{p}_{i-1}$ and $\textbf{p}_i \textbf{p}_{i+1}$.

\subsubsection{Requirement Caused by Extrusion Speed}
Limited by the working principle of material deposition, the allowed extrusion speed has an upper bound $v_{\max}^e$. Therefore, the printer head cannot move too fast. Specifically, we need to have the following constraint
\begin{equation}
    t_i \geq V_i / v_{\max}^e \bar{A}_{i},
\end{equation}
where $V_i$ is the volume of material to be printed between $\mathbf{p}_{i-1}$ and $\mathbf{p}_{i}$, and $\bar{A}_{i}$ is the average cross-sectional area of accumulated material. 

\subsection{Optimization Problem}\label{sec:optProblem}
The trajectory optimization problem by incorporating all the metrics and the constraints discussed above can be formulated as follows.
\begin{align}
    \label{eq:OriginalOptm} &\min_{\{ \bm{\theta}_i \},\{t_i\}} \Phi_{smooth}\\
    \text{subject}& \text{ to } (i=1,\dots,n) \nonumber \\
    \label{cons:ik} & \rev{}{\textbf{q}^A_i=IK^{\mu}(\bm{\omega}_i, \textbf{p}^b_{t,i})} \\
    \label{cons:q} & \textbf{q}_{\min} \leq \textbf{q}_i \leq \textbf{q}_{\max}\\
    \label{cons:vaj} & |\textbf{v}_i| \leq \textbf{v}_{\max}, 
    \  |\textbf{a}_i| \leq \textbf{a}_{\max},
    \  |\textbf{j}_i| \leq \textbf{j}_{\max} \\
    \label{cons:gamma} & \Gamma (\textbf{q}_i) < 0\\
    \label{cons:nNozzle} & \textbf{n}^A_i (\bm{\omega}_i) \cdot \textbf{g} \geq cos(\alpha)\\
    \label{cons:nNormal} & \textbf{n}^B_i (\textbf{q}^B_i) \cdot \textbf{g} \leq -cos(\beta)\\
    \label{cons:nNN} &  \textbf{n}^A_i (\bm{\omega}_i) \cdot \textbf{n}^B_i (\textbf{q}^B_i) \leq -cos(\gamma)\\
    \label{cons:tooltip} & v^t_i \leq v^t_{\max},
    \  |a^t_i| \leq a^t_{\max},
    \  |a^n_i| \leq a^n_{\max}\\
    \label{cons:t} & t_i \geq V_i/v_{\max}^e \bar{A}_{i} \\
    \label{cons:time} &  \Phi_{tm} \leq t^{u}
\end{align}
where $\bm{\theta}_i=[ \bm{\omega}_i^T,(\textbf{q}^B_i)^T ]^T$. \rev{}{$IK^{\mu}(\cdot)$ represents the IK of Robot A according to the $\mu$-th joint configuration, which is selected from the eight possible configurations as stated in Sec.~\ref{sec:initial}.} $\textbf{p}^b_{t,i}$ is determined by $\textbf{q}^B_i$ according to Eqs.\eqref{eq:kmP} and \eqref{eq:kmCons}. Equations \eqref{cons:q}-\eqref{cons:vaj} represent the limits on joint angles, velocities, accelerations, and jerks. Equation \eqref{cons:gamma} is the collision-free constraint, and $\Gamma(\cdot)$ is a proxy collision indication function trained by learning-based methods \cite{Dai2020,Das2020TRO}. Equations \eqref{cons:nNozzle}-\eqref{cons:t} represent the special constraints for additive manufacturing discussed in Sec.~\ref{sec:manuRequirement}, which needs to be replaced by other constraints when subtractive manufacturing such as milling is considered. Lastly, Eq.\eqref{cons:time} gives a requirement on the maximally allowed manufacturing time. 
\section{Numerical Computation} \label{Sec:TrajOpt}
The formulation of trajectory planning proposed above is a large-scale, highly nonlinear, and nonconvex optimization problem. Solving it with off-the-shelf solvers is challenging and expensive in both the computing time and the memory consumption. In order to solve the problem effectively and efficiently, a new scheme of numerical computation is developed by exploiting the special structure of the problem and solving the problem in an out-of-core manner. There are four major steps of our method: initialization, decomposition, sub-problem solving, and result correction. 

\subsection{Initialization} \label{sec:initial}
A good initialization is very important to the convergence speed of numerical computation. When handling the trajectory planning problem on a dual-robot system with 6+2 DoFs, \rev{}{the following steps are taken to determine the configuration index $\mu$ and the initial guess of $\{ \bm{\theta}_{i}=[ \bm{\omega}_i^T,(\textbf{q}^B_i)^T ]^T\}$.}
\begin{enumerate}
\item{We first set $\textbf{n}_i^B=-\textbf{g}$ for all $i \in \{1,\cdots,n\}$ and compute the vector of joint angles $\textbf{q}^B_i$ by the IK of Robot B (i.e., Eq.\eqref{eq:kmP}) using the analytical solution \cite{Yang2013MTM}.}

\item{We then assign $\textbf{n}_i^A=\textbf{g}$ for all $i \in \{1,\cdots,n\}$, which means assigning the third column of $\textbf{R}^b_t$ as $\textbf{g}$ for Robot A employed in Eq.\eqref{eq:kmR}.}

\item{\rev{}{To determine $\textbf{R}^b_t$ and the joint angles $\{ \bm{\theta}_{i}\}$, we need to know the rotation angle $\eta_i$ of the tool around its axis (i.e., $\textbf{n}_i^A$) and the configuration index $\mu$. The graph-based optimization proposed in \cite{Peng2020SciCN,Dai2020, Zhang2021RAL} is employed here to obtain these parameters.}}

\item{To make sure the initial solution is collision-free, collision detection is conducted at every waypoint by using the 
the flexible collision library (FLC) proposed in \cite{FCL}. If a collision occurs at $\textbf{p}_i$, $\eta_i$ will be modified to ensure no collision while minimizing the change of $\eta_i$ (ref.~\cite{Zhang2021RAL}). After that, $\bm{\omega}_i$ is determined by $[{\bm{\omega}}_i] = log (\textbf{R}^b_t)$.}
\end{enumerate}
We denote the initial guess of \rev{}{$\{ \bm{\theta}_i \}$} as $\{ \bm{\theta}_i^0 \}$, which represents the initial configuration of the robotic system. 

For the initial guess of time-sequence, a reasonable choice is based on the angular distance between discrete IK solutions as
\begin{equation} \label{eq:ti0}
    t_i^0 = \max_{j=1,...,8} \{s_i/v_{ini}^t, V_i/v_{\max}^{e} \bar{A}_i, \frac{q_{i,j}-q_{i-1,j}}{v_{max,j}}\},
\end{equation}
where $q_{i,j}$ and $v_{max,j}$ represent the $j$th element of $\textbf{q}_i$ and $\textbf{v}_{max}$. $v_{ini}^t$ should be chosen as a value less than $v_{max}^t$. According to Eqs.\eqref{eq:vt}-\eqref{eq:an}, the initial solution can satisfy the constraints defined in Eq.\eqref{cons:tooltip} as long as $v_{ini}^t$ is small enough. Starting from $v_{ini}^t = v_{max}^t$, we progressively reduce the value of $v_{ini}^t$ until all the constraints in Eq.\eqref{cons:tooltip} are satisfied. The term $(q_{i,j}-q_{i-1,j})/v_{max,j}$ is added in Eq.\eqref{eq:ti0} to indicate the demanded upper bound of joint velocity.

\subsection{Decomposition} \label{sec:ProbDecom}
%
We propose a decomposition based numerical scheme to optimize the trajectory with a large number of waypoints in an out-of-core manner. The strategy is motivated by the block coordinate descent (BCD) method \cite{BCD} and the Benders decomposition \cite{BendersD} but with certain modifications and analysis to fit our optimization framework. While only allowing the variables for a smaller range of waypoints as $\{ \bm{\theta}_i \, | \, i=a,\cdots,b \}$ and $\{ t_i \, | \, i=a+1,\cdots,b \}$ to change during the optimization with $a,b \in \{1,\cdots,n\}$ and $a<b$, the optimization problem becomes
\begin{gather}\label{eq:SubProbOptm}
    \min_{ \substack{\{ \bm{\theta}_i \, | \, i=a,\cdots,b \} \\ \{ t_i \, | \, i=a+1,\cdots,b \} } }  \sum_{i=a-2}^{b+2} \phi_i \Delta s_i\\
    \text{subject to } \text{Eqs.\eqref{cons:ik}-\eqref{cons:time} with } i=a-2,\dots,b+2., \nonumber \\
     \nonumber
\end{gather}
To facilitate the further introduction of this algorithm with different values of $a$ and $b$, the decomposed sub-problem (i.e., Eq.\eqref{eq:SubProbOptm}) is denoted as DS($a,b$) in the rest of this paper. 


\begin{figure} [!t]
\centering
\includegraphics[width=.25\textwidth]{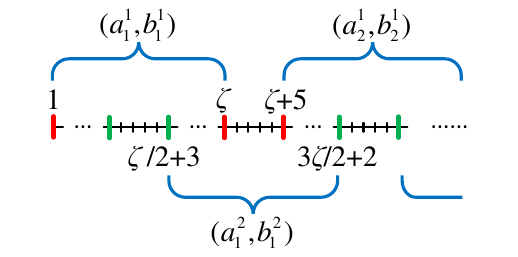}
\caption{An illustration for two sets of toolpath-segments (upper and lower ones), where the toolpath-segments in the same set can be optimized simultaneously in parallel.}
\label{fig:twoSets}
\end{figure}

Analyzing the sub-problem defined in Eq.\eqref{eq:SubProbOptm}, it can be found that DS($a_1,b_1$) and DS($a_2,b_2$) are completely independent when $a_2 - b_1 > 4$. Therefore, they can be solved independently in parallel. Based on this, the decomposition strategy is proposed as follows. 

First, the following two sets of index-pairs are defined to represent two sets of toolpath-segments (see also Fig.\ref{fig:twoSets}). 
\begin{equation}
    \{ (a^1_1,b^1_1), (a^1_2,b^1_2), \cdots, (a^1_{m_1},b^1_{m_1})  \}
\end{equation}
\begin{equation}
    \{ (a^2_1,b^2_1), (a^2_2,b^2_2), \cdots, (a^2_{m_2},b^2_{m_2})  \}
\end{equation}
where $a^1_1=1$, $a^2_1=\zeta/2+3$, $\max\{b^1_{m_1},b^2_{m_2}\}=n$, and $m_1$ and $m_2$ give the sizes of the two sets. For $j=1,2$, we define
\begin{equation}
    a^j_i - b^j_i =\zeta-1, \; b^j_i - a^j_{i-1} = 5.
\end{equation}
$\zeta \geq 4$ is an integer controlling the number of waypoints in each toolpath-segment. In our implementation and all tests, we set $\zeta = 100$ to balance the effectiveness and efficiency. 

Two sets of sub-problems are then defined as
\begin{equation}
    \mathcal{S}_1 = \{ \textnormal{DS}(a^1_1,b^1_1), \textnormal{DS}(a^1_2,b^1_2), \cdots, \textnormal{DS}(a^1_{m_1},b^1_{m_1
    })  \}
\end{equation}
\begin{equation}
    \mathcal{S}_2 = \{ \textnormal{DS}(a^2_1,b^2_1), \textnormal{DS}(a^2_2,b^2_2), \cdots, \textnormal{DS}(a^2_{m_2},b^2_{m_2})  \}
\end{equation}
Based on $\mathcal{S}_1$ and $\mathcal{S}_2$, the decomposed numerical scheme can be presented as Algorithm \ref{alg:algorithm1}, where $\tau$ and $k_{\max}$ are two parameters to control the terminal condition of iterations. $\tau=0.001$ and $k_{\max}=5$ are chosen in our implementation by empirical tuning. Note that the choice of $k_{\max}$ is related to $\zeta$ — a smaller value of $\zeta$ tends to acquire a larger $k_{\max}$. 

\begin{algorithm}[t]
\caption{Decomposition-Based Optimization}\label{alg:algorithm1}
\KwIn{Initial solution $\{ \bm{\theta}_i^0 \}$ and $\{ t_i^0 \}$ }
\KwOut{Optimized solution $\{ \bm{\theta}_i^* \}$ and $\{ t_i^* \}$ }
	
 $\Psi_{old} = \infty$;
 
    Assign $\Psi_{new}$ by the value of the objective function;
    
    $k = 0 $;
 
	\While{ $ (\Psi_{old} - \Psi_{new}) / \Psi_{old} > \tau$ \textbf{\textup{and}} $k < k_{\max}$  }{
        $k = k+1 $;

        $\Psi_{old} = \Phi_{new}$;

        Solve problems in $\mathcal{S}_1$ in parallel;

        Update solution $\{ \bm{\theta}_i \}$ and $\{ t_i \}$;

        Solve problems in $\mathcal{S}_2$ in parallel;

        Update solution $\{ \bm{\theta}_i \}$ and $\{ t_i \}$;

        Assign $\Psi_{new}$ by the value of the objective function;
  
	}

$\{ \bm{\theta}_i^* \} = \{ \bm{\theta}_i \}$ and $\{ t_i^* \} = \{ t_i \}$.

\end{algorithm}

\subsection{Solving Sub-problem} \label{sec:subSolving}
The next issue to be addressed is how to solve the decomposed sub-problem Eq.\eqref{eq:SubProbOptm}. A novel SQP-based algorithm was proposed in \cite{Chen2024RCIM} to solve the smooth trajectory optimization problem in robotic milling. The core of this algorithm is to solve quadratic programming (QP) sub-problems in an iterative way, and it is borrowed to solve the sub-problem in this work. The most critical aspect is how to locally approximate each DS($\cdot$) problem as a QP problem, considering the dual-robot system considered in this paper is more complicated than the 6-DoF single robot system in terms of kinematics. Specifically, we need to approximate the objective function as a quadratic function and approximate all the constraints as linear functions. The approximations are expected to have analytical forms for the efficiency of computation.

For the sake of expression, we denote all the variables in a decomposed sub-problem as a vector $\textbf{x}$ with $\textbf{x}^k$ being its value in the $k$th iteration. The approximation process is presented as follows.

\subsubsection{Increment Vector of Joint Angles} \rev{}{Our approximation is based on local linearization, which requires to compute the first-order differential increments of all parameters w.r.t. the  variables to be optimized. Considering the variables $\{ \bm{\theta}_i=[ \bm{\omega}_i^T,(\textbf{q}^B_i)^T ]^T \}$, the increment of $\textbf{q}_i^A$ can be derived as}
\begin{equation}
    \label{eq:deltaQ}
    \Delta \textbf{q}^A_i =\frac { \partial \textbf{q}^A_i } { \partial \bm{\omega}_i } \Delta \bm{\omega}_i + \frac { \partial \textbf{q}^A_i } { \partial \textbf{q}^B_i } \Delta \textbf{q}^B_i.
\end{equation}
Suppose that $ \bm{\omega}_i = [ \omega_{i,1}, \omega_{i,2}, \omega_{i,3} ]^T $, \rev{}{we can have the following equation for $j \in \{ 1,2,3 \}$ based on the differential kinematics~\cite{Chen2023JMSE,Chen2024RCIM}:}
\begin{equation} \label{eq:qrDw}
    \frac { \partial \textbf{q}^A_i } { \partial \omega_{i,j} } = \textbf{J}^{-1} \mathcal{V}_{\omega_{i,j}},
\end{equation}
where $\textbf{J}$ is the Jacobian of Robot A defined at its tooltip. $ \mathcal{V}_{\omega_{i,j}} = [ (\mathcal{V}^\omega)^T, (\mathcal{V}^v)^T]^T \in \mathbb {R} ^6$ can be considered as the spatial velocity of \{TCS\} w.r.t. to $\omega_{i,j}$, and its matrix form
\begin{equation*}
    [{ \mathcal{V} }_{\omega_{i,j}}] = 
    \begin{bmatrix}
    [{ \mathcal{V} }^\omega] & \mathcal{V}^v  \\
    0 & 0\\
    \end{bmatrix}
\end{equation*}
can be obtained by
\begin{equation}
    [{ \mathcal{V} }_{\omega_{i,j}}] = (\textbf{T}^b_t)^{-1} 
    \begin{bmatrix}
        \partial \textbf{R}^b_t/ \partial \omega_{i,j}  & \textbf{0}  \\
        \textbf{0} & 0
    \end{bmatrix},
\end{equation}
where
\begin{equation} \label{eq:diffRbt}
    \frac{ \partial \textbf{R}^b_t }{ \partial \omega_{i,j}} = \textbf{R}^b_t 
    \left[ \textbf{A}(\bm{\omega}_i) \frac{\partial \bm{\omega}_i}{\partial \omega_{i,j}}  \right],
\end{equation}
\begin{equation}
    \textbf{A}(\bm{\omega}_i)=\textbf{I} - \frac{1- \textup{cos}(\left\| \bm{\omega}_i \right\| )}{\left\| \bm{\omega}_i \right\|^2 }[\bm{\omega}_i]
    +\frac{\left\| \bm{\omega}_i \right\| - \textup{sin}(\left\| \bm{\omega}_i \right\| ) }{\left\| \bm{\omega}_i \right\|^3 }[\bm{\omega}_i]^2.
\end{equation} 
Here $\textbf{I}$ denotes an identity matrix.

Similarly, for any element $q^B_{i,j}, j \in \{1,2\}$ in $\textbf{q}^B_i$, we have 
\begin{equation}
    \label{eq:qrDqp}
    \frac { \partial \textbf{q}^A_i } { \partial q^B_{i,j}  } = \textbf{J}^{-1} \mathcal{V}_{q^B_{i,j}},
\end{equation}
\begin{equation}
    [{ \mathcal{V} }_{q^B_{i,j}}] = (\textbf{T}^b_t)^{-1} 
    \begin{bmatrix}
        \textbf{0}  & \partial \textbf{p}^b_t/ \partial q^B_{i,j}  \\
        \textbf{0} & 0
    \end{bmatrix},
\end{equation}
where $\partial \textbf{p}^b_t/ \partial q^B_{i,j} = \partial \textbf{p}^b_i/ \partial q^B_{i,j}$ has an analytical form according to Eq.\eqref{eq:kmP}.

In summary, $\Delta \textbf{q}^A_i$ has an analytical form based on Eqs.\eqref{eq:deltaQ}, \eqref{eq:qrDw} and \eqref{eq:qrDqp}.

\subsubsection{Increments of Joint Velocity, Acceleration, and Jerk} \rev{}{Based on Eq.\eqref{eq:v_i}, the joint velocity is a function of joint angles and time-sequences. Therefore, its first-order increment is}
\begin{equation}
    \Delta \textbf{v}_i = \sum_{j=i-1}^{i+1}  \frac {\partial \textbf{v}_i} { \partial \textbf{q}_j } \Delta \textbf{q}_j
    + \sum_{j=i}^{i+1} \frac {\partial \textbf{v}_i} { \partial t_j } \Delta t_j,
\end{equation}
where $\Delta \textbf{q}_j = [ (\Delta \textbf{q}^A_j)^T, (\Delta \textbf{q}^B_j)^T ]^T $, $\Delta \textbf{q}^A_j$ has been given in Eq.\eqref{eq:deltaQ}. The increments of joint acceleration and jerk can be obtained in a similar way. All three increments have a concise form as
\begin{equation}\label{eq:deltaVAJ}
    \Delta \textbf{v}_i = \frac{\textup{d} \textbf{v}_i}{\textup{d} \textbf{x}} \Delta \textbf{x}, \;
    \Delta \textbf{a}_i = \frac{\textup{d} \textbf{a}_i}{\textup{d} \textbf{x}} \Delta \textbf{x}, \;
    \Delta \textbf{j}_i = \frac{\textup{d} \textbf{j}_i}{\textup{d} \textbf{x}} \Delta \textbf{x},
\end{equation}
with $\Delta \textbf{x}$ being the increment of the variable vector $\textbf{x}$.

\subsubsection{Local Approximation of the Objective Function} \rev{}{Substituting Eq.\eqref{eq:deltaVAJ} into Eq.\eqref{eq:phi_i} yields the following local approximation of $\phi_i$ near any variable vector $\textbf{x}^k$ as}
\begin{equation}
\begin{aligned}
    \overline{\phi}_i ( \Delta \textbf{x} ) & =
    k_v (\textbf{v}_i+ \frac {\textup{d} \textbf{v}_i} { \textup{d} \textbf{x} } \Delta \textbf{x} )^T \textbf{W} (\textbf{v}_i+ \frac {\textup{d} \textbf{v}_i} { \textup{d} \textbf{x} } \Delta \textbf{x} ) \\
    & + k_a (\textbf{a}_i+ \frac {\textup{d} \textbf{a}_i} { \textup{d} \textbf{x} } \Delta \textbf{x} )^T \textbf{W} (\textbf{a}_i+ \frac {\textup{d} \textbf{a}_i} { \textup{d} \textbf{x} } \Delta \textbf{x} ) \\
    & + k_j (\textbf{j}_i+ \frac {\textup{d} \textbf{j}_i} { \textup{d} \textbf{x} } \Delta \textbf{x} )^T \textbf{W} (\textbf{j}_i+ \frac {\textup{d} \textbf{j}_i} { \textup{d} \textbf{x} } \Delta \textbf{x} )
\end{aligned}
\end{equation}
Thus, the objective function in Eq.\eqref{eq:SubProbOptm} is approximated as
\begin{equation}
    \overline{\Phi}=
    \sum_{i=a-2}^{b+2} \overline{\phi}_i ( \Delta \textbf{x} ) \Delta s_i.
\end{equation}
Since $\overline{\Phi}$ is a quadratic function of $\Delta \textbf{x}$, it has the form
\begin{equation}
    \overline{\Phi}=\Delta \textbf{x}^T \textbf{H} \Delta \textbf{x} + 2 \textbf{f}^T \Delta \textbf{x} + c
\end{equation}
where $\textbf{H}$ is a positive definition matrix, $\textbf{f}$ is a vector, and $c$ is a scalar.

\subsubsection{QP Construction}
Based on the above discussion, the decomposed sub-problem can be analytically approximated near $\textbf{x}^k$ as the following QP-problem.
\begin{align}
    &\min_{\Delta \textbf{x}} \Delta \textbf{x}^T \textbf{H} \Delta \textbf{x} + 2 \textbf{f}^T \Delta \textbf{x}\\
    \text{subject}& \text{ to } (i=a,\dots,b;~ j=a-2,\dots,b+2) \nonumber \\
    & \textbf{q}_{\min} \leq \textbf{q}_i (\textbf{x}^k) + \Delta \textbf{q}_i \leq \textbf{q}_{\max}\\
    & - \textbf{v}_{max} \leq \textbf{v}_j (\textbf{x}^k) + \Delta \textbf{v}_j  \leq \textbf{v}_{\max}\\
    & - \textbf{a}_{max} \leq \textbf{a}_j (\textbf{x}^k) + \Delta \textbf{a}_j  \leq \textbf{a}_{\max}\\
    & - \textbf{j}_{max} \leq \textbf{j}_j (\textbf{x}^k) + \Delta \textbf{j}_j  \leq \textbf{j}_{\max}\\
    & \Gamma (\textbf{x}^k) + \left( {\textup{d} \Gamma}/{\textup{d} \textbf{q}_i} \right) \Delta \textbf{q}_i < 0\\
    & ({\textup{d} \textbf{n}^A_i}/{\textup{d} \bm{\omega}_i}) \Delta \bm{\omega}_i \cdot \textbf{g} \geq \textup{cos}(\alpha) - \textbf{n}^A_i(\textbf{x}^k) \cdot \textbf{g} \\
    & ({\textup{d} \textbf{n}^B_i}/{\textup{d} \textbf{q}^B_i}) \Delta \textbf{q}^B_i \cdot \textbf{g} \leq -\textup{cos}(\beta) - \textbf{n}^B_i(\textbf{x}^k) \cdot \textbf{g} \\
    & \frac{\textup{d} (\textbf{n}^A_i \cdot \textbf{n}^B_i)}{\textup{d} \textbf{x}} \Delta \textbf{x} \leq -\textup{cos}(\gamma) - \textbf{n}^A_i (\textbf{x}^k) \cdot \textbf{n}^B_i (\textbf{x}^k) \\
    & v^t_i (\textbf{x}^k) + \left( {\textup{d} v^t_i}/{\textup{d} \textbf{x}} \right) \Delta \textbf{x} \leq v^t_{\max}\\
    & |a^t_i (\textbf{x}^k) + \left( {\textup{d} a^t_i}/{\textup{d} \textbf{x}} \right) \Delta \textbf{x} | \leq a^t_{\max}\\
    & |a^n_i (\textbf{x}^k) + \left( {\textup{d} a^n_i}/{\textup{d} \textbf{x}} \right) \Delta \textbf{x}  | \leq a^n_{\max}\\
    & t_i (\textbf{x}^k) + \Delta t_i \geq V_i/v_{\max}^{e} \Bar{A}_{i} \\
    & \sum_{i=a+1}^b \Delta t_i \leq t^u - \Phi_{tm} (\textbf{x}^k)
\end{align}
where $\Delta \textbf{q}_i = [(\Delta \textbf{q}^A_i)^T,(\Delta \textbf{q}^B_i)^T]^T$.
$ {\textup{d} \Gamma}/{\textup{d} \textbf{q}_i}$ can be analytical as long as the collision indication function $\Gamma$ is represented in an analytical form. 
Since $\textbf{n}^A_i$ and $\textbf{n}^B_i$ are the third columns of $\textbf{R}^b_t$ and $\textbf{R}^b_i$ respectively, ${\textup{d} \textbf{n}^A_i}/{\textup{d} \bm{\omega}_i}$ and ${\textup{d} \textbf{n}^B_i}/{\textup{d} \textbf{q}^B_i}$ are obtained from Eq.\eqref{eq:diffRbt} and the differentiation of Eq.\eqref{eq:kmP}, thereby allowing the calculation of ${\textup{d} (\textbf{n}^A_i \cdot \textbf{n}^B_i)}/{\textup{d} \textbf{x}}$. The analytical form for ${\textup{d} v^t_i}/{\textup{d} \textbf{x}}$, ${\textup{d} a^t_i}/{\textup{d} \textbf{x}}$, and ${\textup{d} a^n_i}/{\textup{d} \textbf{x}}$ can be obtained by differentiating Eqs. \eqref{eq:vt}-\eqref{eq:an}. 

By the algorithm presented in \cite{Chen2024RCIM}, the solution can be obtained by iteratively solving a series of QP-problems. Note that since the QP-problem is sparse, it can be efficiently solved by using advanced libraries such as OSQP \cite{osqp}.

\subsection{Result Correction}\label{sec:resultCorrection}
This step is conducted after determining the optimized $\{\bm{\theta}_i^* \}$ and $\{ t_i^* \}$. As the collision indication function $\Gamma (\cdot)$ is obtained by the learning-based method that does not present the obstacles precisely, the resultant $\{\bm{\theta}_i^* \}$ cannot completely assure a collision-free machining process. Therefore, we need to verify the result of optimization by using a geometry-based collision detection library. If a collision happens at $\textbf{p}_i$, $\bm{\theta}_i$ is modified by a bi-sectional search method to find a collision-free solution between $\bm{\theta}_i^*$ and $\bm{\theta}_i^0$. 
\section{Implementation Details and Discussion} \label{sec:discussion}
In this section, we discuss the details of collision detection, normalization for different objectives, and the generalization of our approach. 

\subsection{Collision Detection}
For the proxy detector of collision $\Gamma (\cdot)$, there are many learning-based methods to train it. For subtractive manufacturing, only the shape of the target workpiece needs to be considered in training. Differently, the shape of a workpiece changes from layer to layer while accumulating materials, and different proxy functions need to be trained for each layer.

The coarse-to-fine sampling strategy proposed in \cite{Dai2020} is adopted in our implementation to construct a dataset more adaptive to the shape of obstacles. Firstly, we obtain an approximately uniform sampling near obstacles in configuration space (C-space) by randomly sampling on the working layer surface and up-scaling in C-space. After that, the refinement step is conducted to increase the density of samples near the collision boundary in C-space. 

\begin{figure}[!t]
\centering
\includegraphics[width=.4\textwidth]{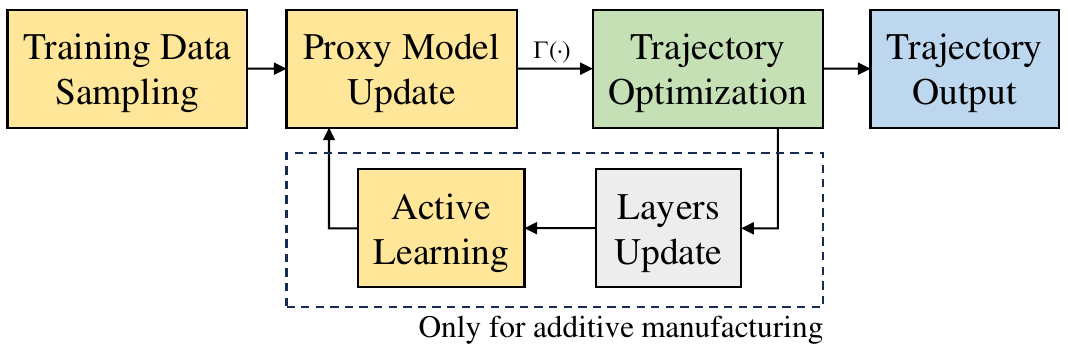}
\caption{The training process of the proxy functions for collision detection.}
\label{fig:colliTrain}
\end{figure}

The learning-based algorithm, Fastron \cite{Das2020TRO}, is used to train the proxy function $\Gamma (\cdot)$. For the first printing layer, the proxy model is trained based on the samples obtained by the above method. At the same time, we get a set of supporting samples $\mathcal{S}_1$ for this proxy model. For the $i$th layer with $i>1$, the Fastron Active Learning Algorithm proposed in \cite{Das2020TRO} helps get a new set of samples $\mathcal{A}_i$ based on the supporting samples $\mathcal{S}_{i-1}$ of the last proxy model. Then the proxy model is updated based on samples $\mathcal{S}_{i-1} \cup \mathcal{A}_i$ and the current proxy model. In this way, we do not need to rebuild a new function for each layer but to update the existing one. The training flowchart for learning collision detection proxy functions is shown in Fig.~\ref{fig:colliTrain}. 

In special cases where the surface of a workpiece is relatively flat, the probability of a collision occurring is very low, especially when $\textbf{n}^A_i$ and $\textbf{n}^B_i$ are constrained. Therefore, the collision-free constraint in Eq.\eqref{cons:gamma} can be removed from the optimization problem to accelerate the computation. This also avoids the steps of dataset generation and proxy model training and further reduces the computation time. It should be noted that the resultant trajectory is still ensured to be collision-free thanks to the result correction process presented in Sec.~\ref{sec:resultCorrection}.

\subsection{Normalization}\label{subsecNorm}
To facilitate the value setting of $k_v$, $k_a$, and $k_j$ in Eq.\eqref{eq:phi_i}, it is better to normalize the corresponding terms in the kinematic metric. The normalization method proposed in \cite{Chen2024RCIM} is conducted in all our tests. For example, if we denote $\textbf{v}_i^T \textbf{W} \textbf{v}_i$ in Eq.\eqref{eq:phi_i} by $\phi_i^{Vec}$, the normalized result of $\phi_i^{Vec}$ is 
\begin{equation}
    \label{eq:normPhiVel}
    \Tilde{\phi}_i^{Vec}=\frac{\phi_i^{Vec} - \phi_{min}^{Vec}}{\phi_{max}^{Vec} - \phi_{min}^{Vec}}
\end{equation}
where $\phi_{max}^{Vec}$ and $\phi_{min}^{Vec}$ are the maximum and minimum values of $\phi_i^{Vec}$ for $i=1,\dots,n$ based on the initial solution. 

Similarly, the terms of $\textbf{a}_i^T \textbf{W} \textbf{a}_i$ and $\textbf{j}_i^T \textbf{W} \textbf{j}_i$ in Eq.\eqref{eq:phi_i} are normalized in the same way. After normalization, we choose the weight values as $k_v = 0.1$, $k_a = 0.5$, and $k_j = 1$ by experiments and employ them in all examples tested in this paper.

\subsection{Generalization} \label{Sec:DiscuGenera}
Our formulation is general which can handle the optimization of joint poses and the time-sequence in both a simultaneous and a decoupled manners. Specifically, we can fix the value of time-sequence $\{t_i\}$ so that the speed of tooltip's motion can be controlled precisely. On the other aspect, when there are special requirements of joint poses on different waypoints, we can compute and determine the poses $\{\bm{\theta}_i\}$ first and then only optimize the time-sequence $\{t_i\}$ by our optimization framework. Moreover, we can replace the time-sequence by the sequence of arc-length parameters $\{s_i\}$ for waypoints on the toolpath. As a result, the objective function becomes a metric evaluating the geometric smoothness of joint paths (ref.~\cite{Chen2023JMSE,Lu2021RCIM}).

As a general framework of trajectory optimization using kinematic redundancy, our method can also be applied to a single-robot system. For the system consisting of a 6-DoF robot and a fixed printer head shown in Fig.~\ref{fig:newRob}, \{WCS\} is located at the end of the robot. The kinematic model is
\begin{equation}
    \label{kmSingle}
    \textbf{T}^b_w(\textbf{q}_i) \textbf{T}^w_{l_i} \textbf{T}^{l_i}_t
    = \textbf{T}^b_t
\end{equation}
with
\begin{equation}
    \textbf{T}^{l_i}_t = 
    \begin{bmatrix}
        \textbf{R}^i_t & \textbf{0} \\
        \textbf{0} & 1
    \end{bmatrix}
\end{equation}
The exponential coordinate of $\textbf{R}^i_t$ can be chosen as the independent variable since $\textbf{T}^w_{l_i}$ and $\textbf{T}^b_t$ are constant. Replacing $\{\bm{\theta}_i\}$ in the optimization problem with this new independent variable yields the trajectory optimization model for the single robot system, and the method in Sec. \ref{Sec:TrajOpt} can be directly used to solve it. An example of such a single-robot system with kinematic redundancy is also given in the following section.

%
%
\section{Simulations and Experiments}\label{sec:validation}
\begin{figure}[!t]
\centering
\includegraphics[width=\linewidth]{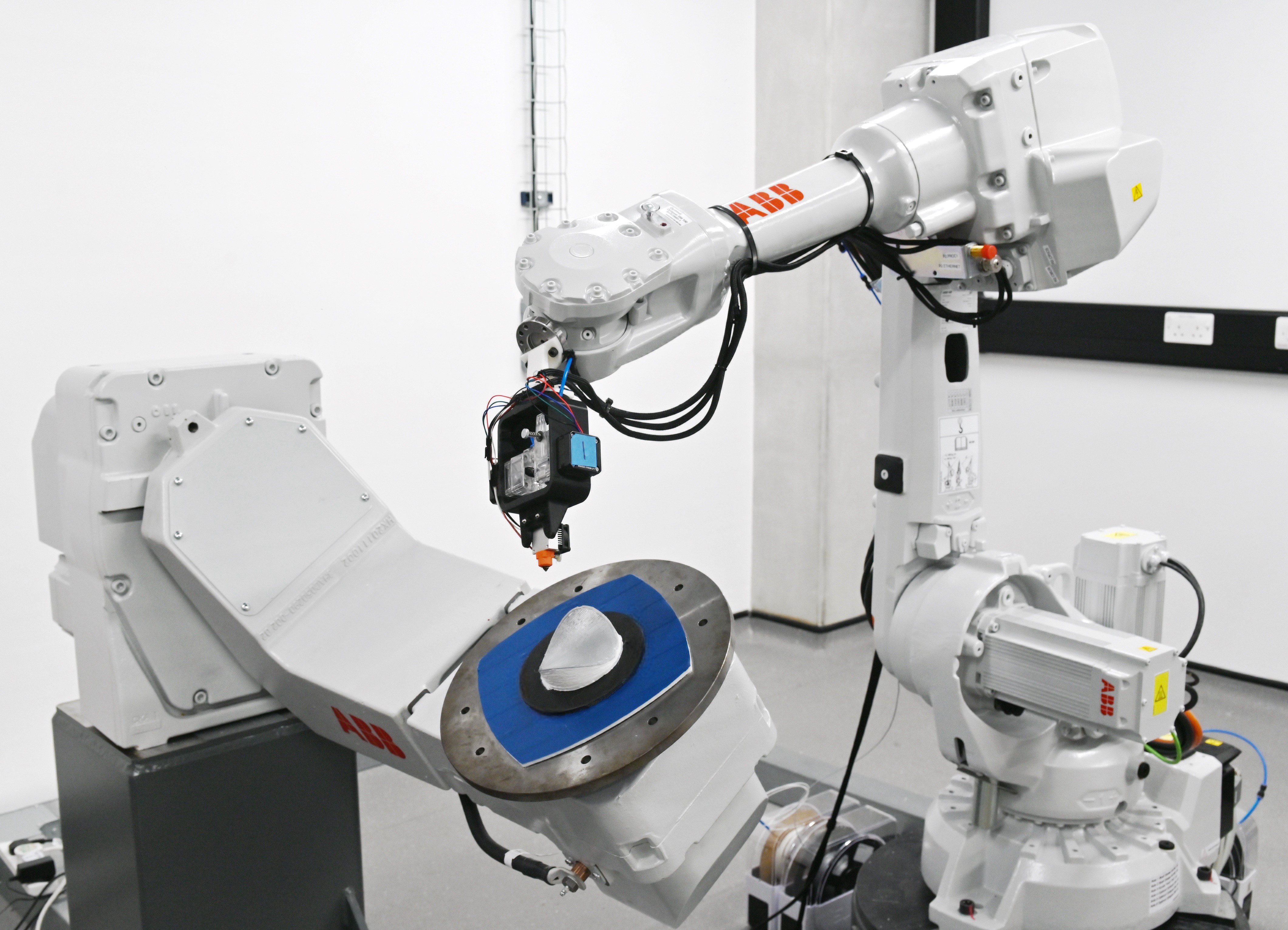}
\caption{The dual-robot system used for AM experiments, where the printer head is mounted on the end-effector of the 6-DoF robot and the workpiece is mounted on the working plate of a position table with 2-DoF.}\label{fig:ExSetUp}
\end{figure}

In this section, simulations and physical fabrication experiments of additive manufacturing are conducted to validate the effectiveness of our method. The robot system used consists of a robot with 6-DoF (ABB IRB-2600) and a position table with 2-DoF (ABB IRBP-A), as shown in Fig.~\ref{fig:ExSetUp}. All programs are implemented by C++ and tested on a PC with an Intel Core i9 CPU at 3GHz and 32GB RAM.
The method presented in this paper has been conducted to optimize trajectories for curved layers in AM as freeform surfaces, the results of which are discussed below and can also be found in the supplementary video: \url{https://youtu.be/vILrYwFufUk}. \rev{}{We have made our source code publicly accessible at the GitHub site: \url{https://github.com/Yongxue-Chen/ConcurrentTrajOpt}.} 

\subsection{Example I}
\begin{figure}[!t]
\centering
\includegraphics[width=.35\textwidth]{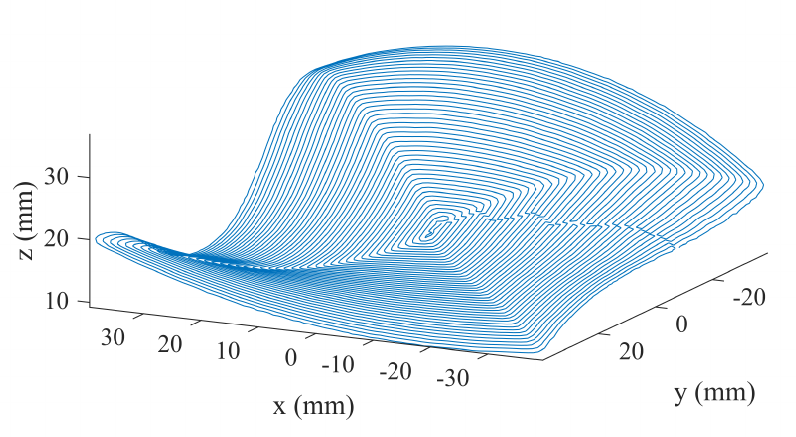}
\vspace{-5pt}
\caption{The toolpath of a curved layer for AM employed in Example I.}
\label{fig:Ex1Path}
\end{figure}
The first example is a curved layer with $7,497$ waypoints as shown in Fig. \ref{fig:Ex1Path}. The parameters for orientation constraints are set to $\alpha=20^{\circ}$, $\beta=8^{\circ}$, $\gamma=12^{\circ}$. The upper bound $t^u$ is set as the manufacturing time of the initial solution. 

\begin{figure} [!t]
\centering
\includegraphics[width=.485\textwidth]{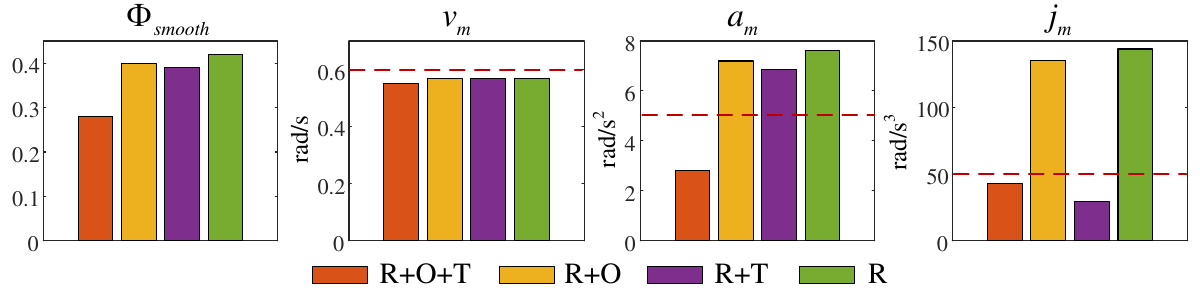}
\caption{Ablation study by using Example I, where $v_m$, $a_m$, and $j_m$ are the maximum values of joint velocity, acceleration, and jerk on the resultant trajectories. The dashed lines represent the maximally allowed joint velocity, acceleration, and jerk as constraints imposed in optimization.}
\label{fig:Ex1SolCom}
\end{figure}

\begin{table}[!t]
\begin{center}
\caption{Optimization results of Example I: Ablation Study.}
\label{Table:solution1}
    
\begin{tabular}{|l|r|r|r|r|}
\hline
 & $\Phi_{smooth}$ & $v_m$ & $a_m$ & $j_m$\\
\hline
\hline
Before Optimization & 4.48   & 1.22   &  27.06 & 812.24 \\ 
\hline
R+O+T &0.28 &0.55 &2.82 &43.12\\
R+O   & 0.40   & 0.57 &7.20 &135.55\\ 
R+T   & 0.39   & 0.57 &6.87 &29.39\\  
R only  &  0.42   & 0.57 & 7.61 & 143.98\\  
\hline
Maximally Allowed &    & 0.60 &5.00 &50.00\\  
\hline
\end{tabular}

\end{center}
\end{table}

\begin{figure}[!t]
\centering
\includegraphics[width=.4\textwidth]{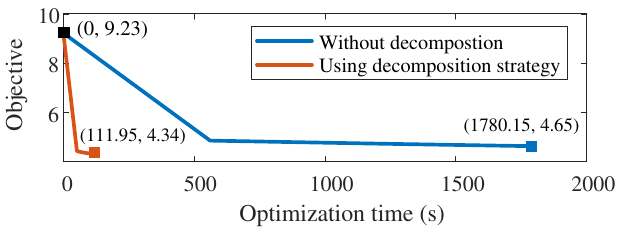}
\vspace{-5pt}
\caption{Effectiveness study of the decomposition scheme by Example I.}
\label{fig:Ex1Decom}
\end{figure}

\subsubsection{Importance of concurrent optimization}
For trajectory optimization in robot-assisted AM, the most common framework is to just optimize the robot redundancy with predefined tool orientation and tooltip speed (e.g., \cite{Dai2020}). We conduct an ablation study to demonstrate the importance of concurrent optimization as follows.

Using `R', `O' and `T' to denote the aspects of redundancy, tool-orientation and time-sequence in optimization respectively, the result of concurrent optimization is represented as `R+O+T' in the comparisons given in Fig.~\ref{fig:Ex1SolCom} and Table \ref{Table:solution1}. The result of only optimizing the kinematic redundancy is given as `R'. 
The effectiveness of different aspects is also shown by fixing the tool-orientaions (i.e., the `R+T' result) or fixing the time-sequence (i.e., the `R+O' result). Besides of the smoothness metric $\Phi_{smooth}$, we also check the maximal joint velocity (denoted by $v_m$), the maximal joint acceleration (denoted by $a_m$), and the maximal joint jerk (denoted by $j_m$) throughout the optimized trajectory among all components. They are compared with the maximally allowed values in Table \ref{Table:solution1} and Fig.~\ref{fig:Ex1SolCom} (visualized as dash lines).

The comparison shows that concurrent optimization has the best performance. It reduces the objective function value by 93.75\% with all constraints of maximal values satisfied.
In contrast, the other three solutions do not meet the constraints on joint acceleration and/or jerk. This study proves the importance of our concurrent optimization framework.

\subsubsection{Effectiveness of decomposition scheme}
The effectiveness of the decomposition scheme proposed in Sec.~\ref{sec:ProbDecom} is tested on this example. For this path containing $7,497$ waypoints, directly optimizing the entire path without the decomposition failed on our PC due to the out-of-memory reason. To compare the performance, tests are conducted by only using the first $1,000$ waypoints of the toolpath. The optimization process is as shown in Fig. \ref{fig:Ex1Decom}. It can be found that the decomposition scheme can reduce the computing time by more than 93\% without compromising the quality of optimization. 

\begin{figure}[!t]
\centering
\includegraphics[width=.4\textwidth]{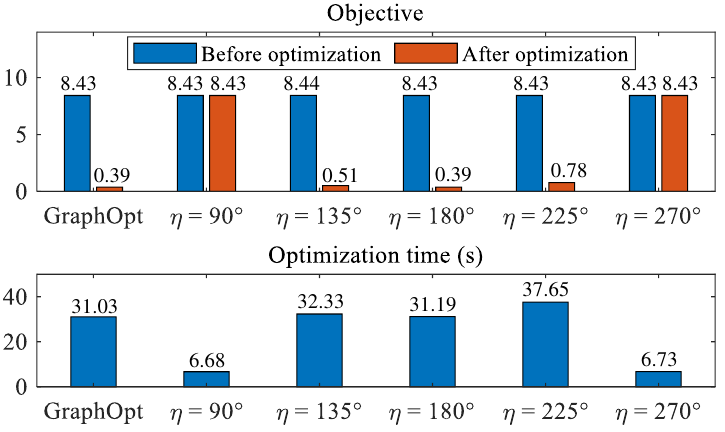}
\vspace{-5pt}
\caption{\rev{}{Sensitivity study of initial solutions in Example I, where `GraphOpt' represents the initial solution obtained from graph-based optimization. The others are derived by setting the redundant angle $\eta$ to different fixed values.}}
\label{fig:Ex1Sensi}
\end{figure}
\subsubsection{Sensitivity study} \rev{}{Now we test the sensitivity of our optimization method when using different initial solutions for the first $500$ waypoints of a toolpath (i.e., Example I). The results are presented in Fig.~\ref{fig:Ex1Sensi}. The results indicate that when the initial solution lies within a certain range, both the optimization time and the final results are similar, demonstrating the robustness of our algorithm. However, due to the characteristics of numerical optimization methods, the algorithm does not converge to the optimal solution when the initial solution deviates significantly from the optimum -- e.g., by setting the redundant angle $\eta=90^\circ$ or $\eta=270^\circ$. This test also validates the effectiveness of the initialization scheme as proposed in Sec.~\ref{sec:initial} -- indicated as `GraphOpt' in Fig.~\ref{fig:Ex1Sensi}, which gives the best performance among other initial solutions.}

\begin{figure*}[!t]
    \centering
    \includegraphics[width=.99\textwidth]{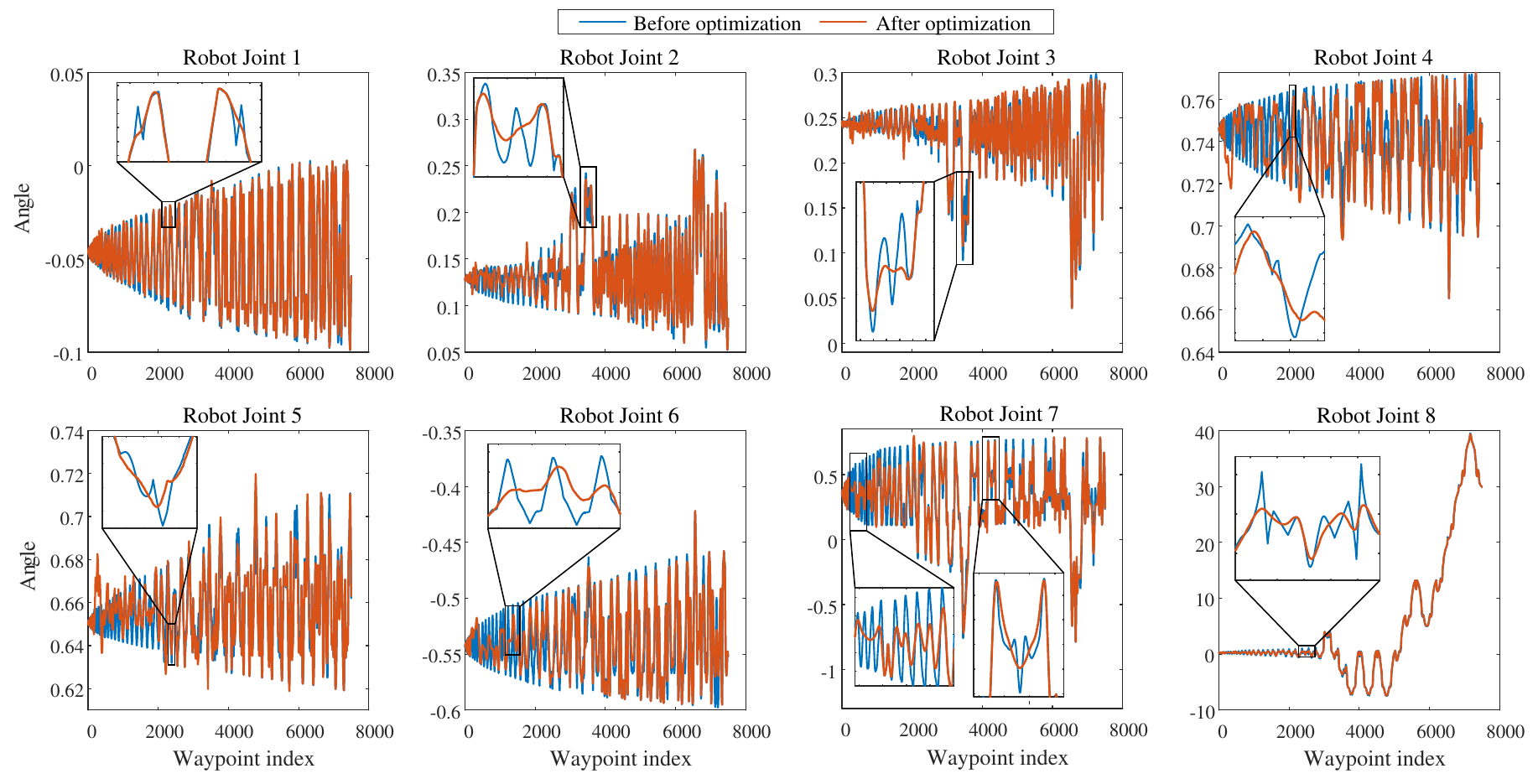}
    \vspace{-5pt}
    \caption{The robot joint path of Example I, where the joints 1-6 are from Robot A and the joints 7 \& 8 belong to Robot B. The unit of joint angles is $rad$.}
    \label{fig:Ex1Joints}
\end{figure*}
\begin{figure} [!t]
\centering
\includegraphics[width=.46\textwidth]{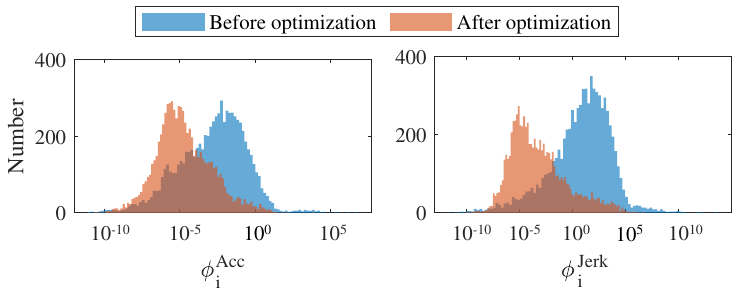}
\vspace{-5pt}
\caption{Histograms of smoothness metrics $\phi_i^{Acc}$ and $\phi_i^{Jerk}$ in Example I.}
\label{fig:Ex1DistriAJ}
\end{figure}

\subsubsection{Results and physical verification}
The results of optimization on all joint paths are shown in Fig.~\ref{fig:Ex1Joints}. It can be observed that the fluctuations in joint movements can be significantly reduced. This is mainly due to the improvement of kinematic smoothness. To further study the performance of our trajectory optimization approach, we analyze the histograms of two metrics $\phi_i^{Acc}= \textbf{a}_i^T \textbf{W} \textbf{a}_i$ and $\phi_i^{Jerk}= \textbf{j}_i^T \textbf{W} \textbf{j}_i$ that give the kinematic smoothness at the $i$-th waypoint. The histograms before vs. after optimization are as shown in Fig.\ref{fig:Ex1DistriAJ}, where the histograms after optimization are shifted substantially to the left, indicating that the smoothness at the vast majority of the waypoints has been improved.

    





\begin{table}
\begin{center}
\caption{Acceleration measured for Example I.}
\label{Table:VibrationEx1}
\begin{tabular}{|c||c|c||c|c|}
\hline
& \multicolumn{2}{c||}{Robot A} & \multicolumn{2}{c|}{Robot B} \\
\cline{2-5}
 & Maximum & Average & Maximum & Average \\
\hline
\hline
Before optimization    & 0.1140   & 0.0066 & 0.1939 & 0.0441\\   
After optimization    & 0.1040   & 0.0058 & 0.1869 & 0.0353\\
\hline
\end{tabular}
\begin{flushleft}
\item$^\dag$All values are given in terms of the gravitational acceleration $g$.
\end{flushleft} 
\end{center}
\end{table}

\begin{figure}[!t]
\centering
\includegraphics[width=.46\textwidth]{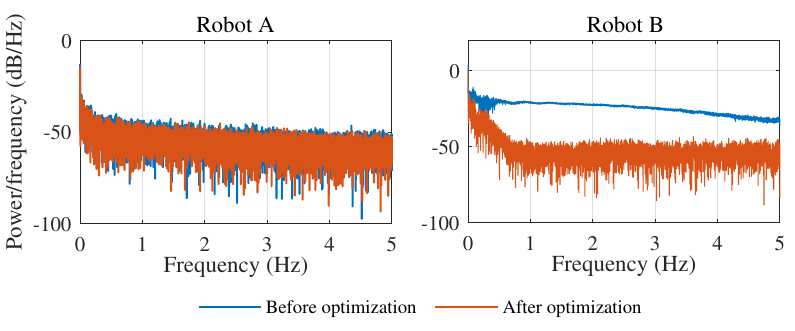}
\vspace{-10pt}
\caption{The periodogram power spectral density estimate of the measured acceleration in Example I.}
\label{fig:Ex1AccSensor}
\end{figure}

The performance of our trajectory optimization approach has been verified via physical fabrication conducted on a dual-robot system as shown in Fig.~\ref{fig:ExSetUp}. Vibrations at the end-effectors of robots A and B are measured by accelerometers, and the results are illustrated in Table \ref{Table:VibrationEx1} and Fig.~\ref{fig:Ex1AccSensor}. After optimizing the trajectory, the average values of accelerations are reduced by 12.12\% and 19.95\% on robots A and B respectively. The periodogram power spectral density estimate of the measured data is shown in Fig. \ref{fig:Ex1AccSensor}. The improvement in the vibration of Robot B is significant, as evidenced by a lower power spectral density.

\begin{figure}[!t]
\centering
\includegraphics[width=.5\textwidth]{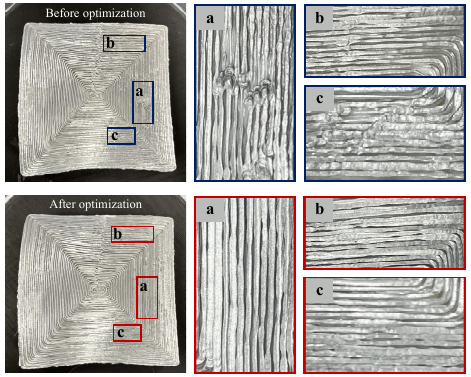}
\caption{Results of physical fabrication for Example I, from which we can find significant improvement after optimization.}
\label{fig:Ex1Exp}
\end{figure}

We also compared the quality of 3D printed curved layers by using both the trajectories before and after optimization -- see the results shown in Fig. \ref{fig:Ex1Exp}. When using the unoptimized trajectory, it is difficult to synchronize the speed of material extrusion with the nozzle movement due to the jerky motion of robotic joints. As a result, the extruded material is quite uneven in regions such as highlighted Areas `b' and `c'. In Area `a', the situation is even worse -- i.e., the material breaks and piles up. All these problems can be clearly improved to achieve better surface quality by using the optimized trajectory. 

\begin{figure}[!t]
    \centering
    \includegraphics[width=.25\textwidth]{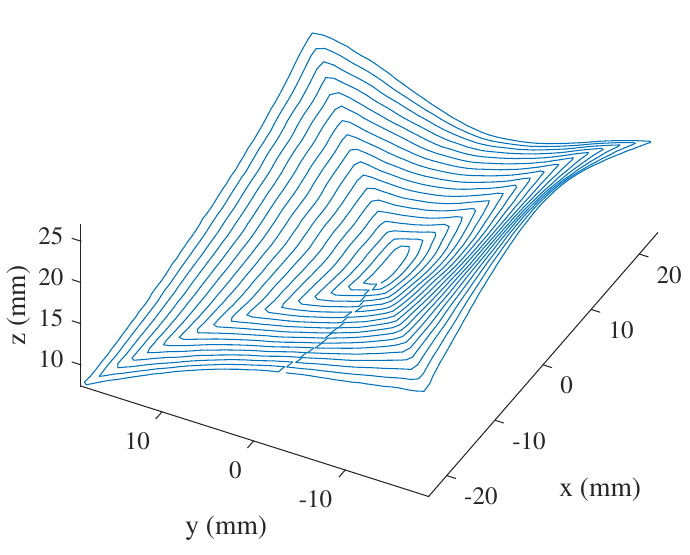}
    \caption{The toolpath of a curved layer for AM employed in Example II.}
    \label{fig:Ex2Path}
\end{figure}

\begin{figure*}[!t]
\centering
\includegraphics[width=.99\textwidth]{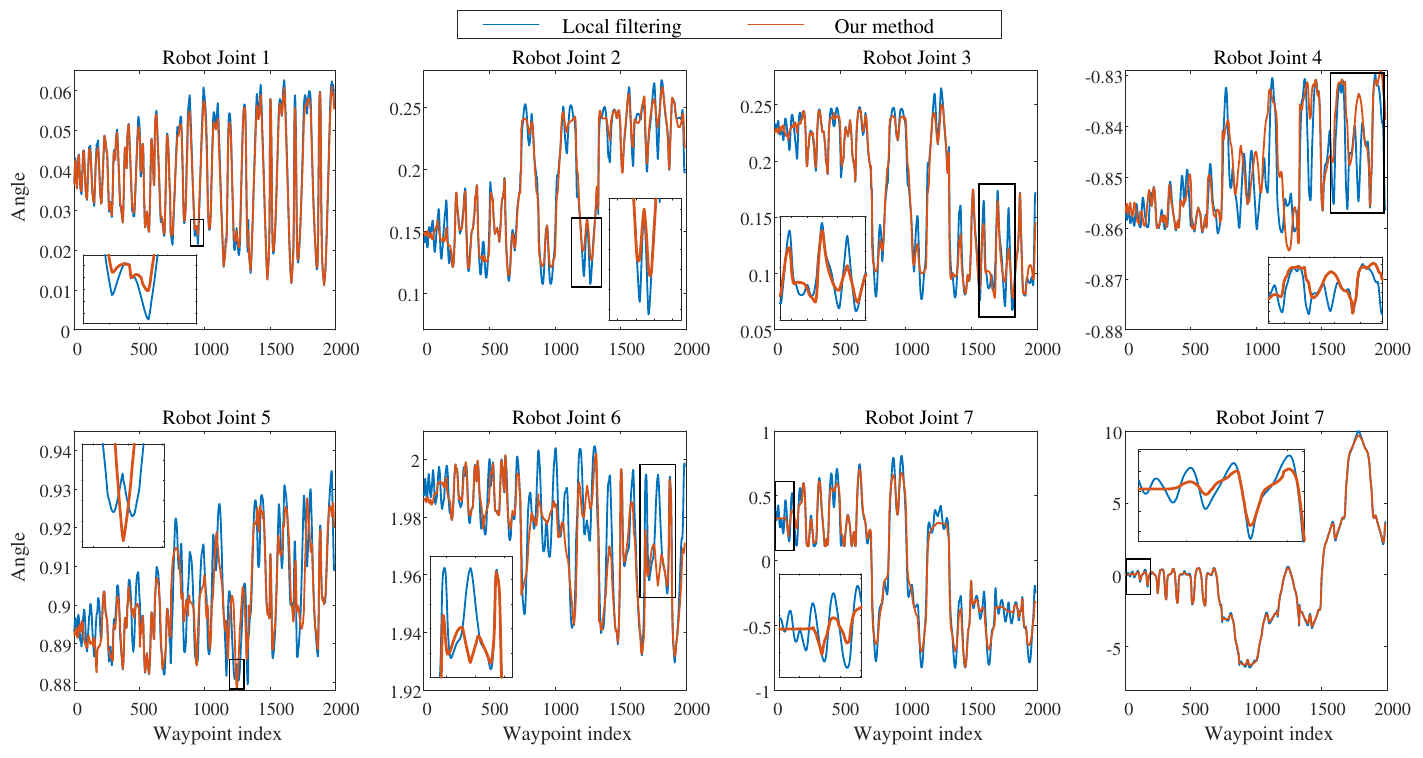}
\vspace{-5pt}
\caption{The robot joint path of Example II, where the joints 1-6 are from Robot A and the joints 7 \& 8 belong to Robot B. The unit of joint angles is $rad$.}
\label{fig:Ex2Joints}
\end{figure*}

\subsection{Example II}
The toolpath of Example II contains of $1,987$ waypoints as shown in Fig. \ref{fig:Ex2Path}. The comparison with the local filtering method presented in \cite{Dai2020} is conducted. The local filtering method needs a prescribed time-sequence, and the optimization object is to minimize the joint jerk. The maximum iteration time of the local filtering method is set to 120. For the purpose of comparison, the time-sequence in our method is also fixed during optimization (i.e., the `R+O' scheme is taken). In both methods, the allowable maximum velocity, acceleration, and jerk are set to $0.5rad/s$, $8rad/s^2$, and $60rad/s^3$. The joint paths obtained by \cite{Dai2020} and ours are given in Fig.~\ref{fig:Ex2Joints}, where less joint shaking is observed in our result.

\begin{figure}[!t]
    \centering
    \includegraphics[width=.43\textwidth]{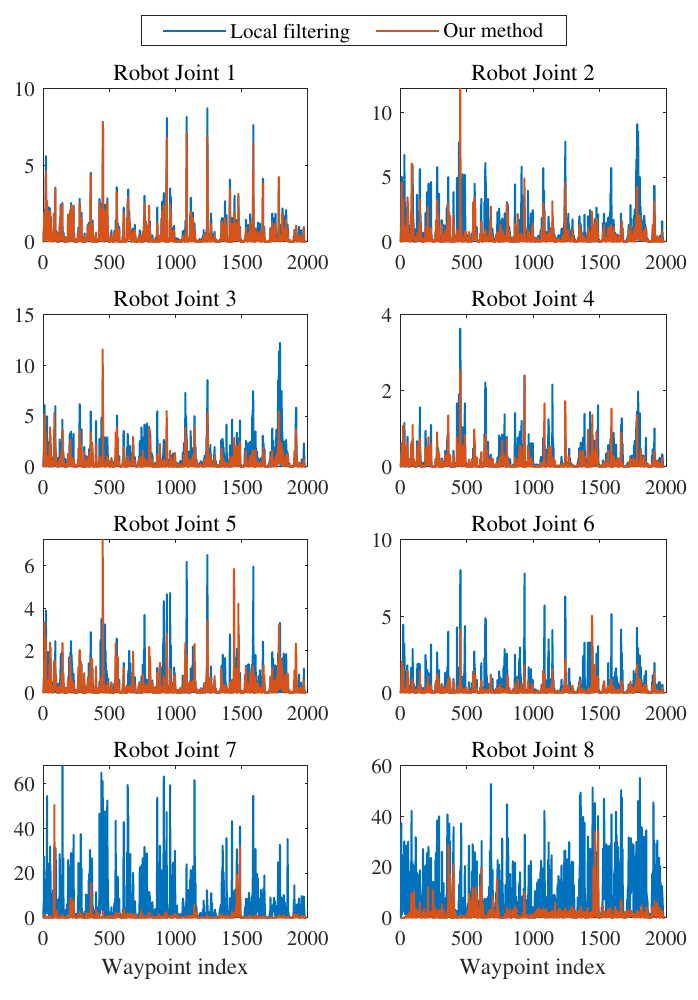}
    \vspace{-8pt}
    \caption{The comparison for the absolute values of jerks on different joints in Example II.}
    \label{fig:Ex2jointJerk}
\end{figure}

\begin{table}[!t]
\caption{Optimization results in Example II.}
\label{Table:solution2}
\begin{tabular}{|c|r|r|r|r|}
\hline
\multicolumn{2}{|c|}{} & Before Opt. & Our method &  Local filter \\
\hline
\multicolumn{2}{|c|}{Computing Time (sec.)} & & 81.10 & 225.36 \\
\hline
\hline
 & Joint 1    & 0.46 & 0.36 & 0.46  \\   
   & Joint 2    & 0.67 & 0.37 & 0.67 \\
       & Joint 3    & 0.75 & 0.42 & 0.74 \\
Average        & Joint 4    & 0.16 & 0.10 & 0.16 \\ 
Jerk         & Joint 5    & 0.38 & 0.27 & 0.38 \\
        & Joint 6    & 0.41 & 0.17 & 0.41 \\ 
        & Joint 7    & 5.21 & 0.60 & 5.21 \\
        & Joint 8    & 9.47 & 1.50 & 7.99 \\
\hline
      & Joint 1    & 8.72 & 7.81 & 8.72 \\   
       & Joint 2    & 9.09 & 11.83 & 9.09 \\
 & Joint 3    & 12.25 & 11.58 & 12.21 \\
Maximal   & Joint 4    & 3.06 & 2.54 & 3.63 \\ 
Jerk        & Joint 5    & 6.52 & 7.24 & 6.52 \\
        & Joint 6    & 8.51 & 5.04 & 8.02 \\ 
        & Joint 7    & 67.99 & 50.42 & 68.00 \\
        & Joint 8    & 182.83 &34.40 & 55.11 \\
\hline
\hline
Average & Robot A    & 0.0094 & 0.0060 & 0.0094\\
Velocity & Robot B & 0.2171 & 0.1514 & 0.2163\\
\hline
Maximal & Robot A     & 0.0877 & 0.0387 & 0.0876 \\
Velocity & Robot B & 0.5000 & 0.5000 & 0.5000\\
\hline\hline
Average & Robot A    & 0.0392 & 0.0236 & 0.0392 \\
Acceleration & Robot B & 0.6802 & 0.1917 & 0.6802\\
\hline
Maximal & Robot A    &  0.9700 & 0.5743 & 0.9696\\
Acceleration & Robot B & 8.7373 & 3.8540 & 5.0408\\
\hline
\end{tabular}
\begin{flushleft}
\item$^\dag$The joints 1-6 are from Robot A and the joints 7 \& 8 belong to Robot B.
\end{flushleft} 
\vspace{-10pt}
\end{table}

The comparison of joint jerks has also been given in Fig. \ref{fig:Ex2jointJerk} and Table \ref{Table:solution2}. The trajectory obtained by our method has a lower average jerk in general, while those obtained by local filtering only perform better at some local maxima. The other observation to note is that the maximum jerk of Joint 7 (i.e., the first joint of Robot B) obtained by local filtering is still larger than the maximally allowed value $60~rad/s^3$ after 120 iterations. We can also find from Table \ref{Table:solution2} that both the average and the maximal values of joint velocity and acceleration have been reduced by our method as they are incorporated in our formulation as smoothness metrics. Differently, they are nearly not changed by the local filtering method \cite{Dai2020} as not included in the objective function of optimization. 

\begin{figure}[!t]
    \centering
    \includegraphics[width=.38\textwidth]{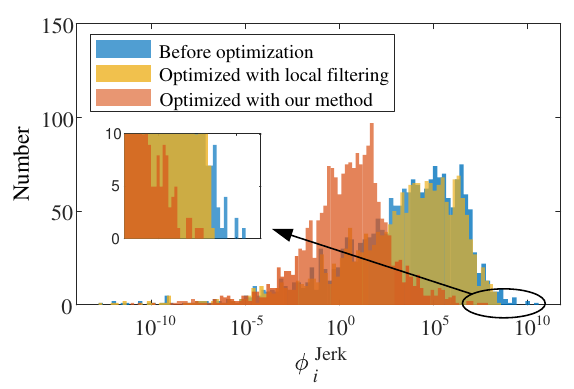}
    \vspace{-8pt}
    \caption{Histograms of smoothness metrics $\phi_i^{Jerk}$ for Example II.}
    \label{fig:Ex2PhiJerk}
\end{figure}

Fig. \ref{fig:Ex2PhiJerk} illustrates that the histogram of the smoothness metrics $\phi_i^{jerk}$ obtained by both the local filtering method and our method. Our method can significantly shift the whole distribution to the left side while the improvement given by the local filtering method is mainly reflected at the right border of the distribution. This is because the local filtering method only conducts optimization for local parts with maximal jerk values.

\begin{figure}[!t]
\centering
\includegraphics[width=.4\textwidth]{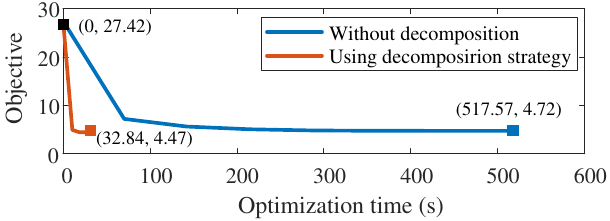}
\vspace{-5pt}
\caption{Effectiveness study of the decomposition scheme by Example II.}
\label{fig:Ex2Decom}
\end{figure}

Compared with the local filtering method, the computational efficiency of our method has been improved tremendously -- i.e., the computing time has been reduced by 64.01\% from 225.36~sec. to 81.10~sec. The effectiveness of the decomposition scheme is also verified in this example. Again, we take the first 500 waypoints for testing because of the limit of computer memory. The result shown in Fig. \ref{fig:Ex2Decom} illustrates that the computing time is reduced by 93.65\% with the help of the decomposition scheme.

\begin{table}
\begin{center}
\caption{Acceleration measured in Example II.}
\label{Table:VibrationEx2}
\begin{tabular}{|c||c|c||c|c|}
\hline
& \multicolumn{2}{c||}{Robot A} & \multicolumn{2}{c|}{Robot B} \\
\hline
 & Maximum & Average & Maximum & Average \\
\hline
\hline
Local filtering    & 0.0661   & 0.0077 & 0.1493 & 0.0239\\   
Proposed method    & 0.0593   & 0.0067 & 0.0870 & 0.0165\\
\hline
\end{tabular}
\begin{flushleft}
\item$^\dag$All values are given in terms of the gravitational acceleration $g$.
\end{flushleft} 
\end{center}
\end{table}

\begin{figure} [!t]
    \centering
    \includegraphics[width=.47\textwidth]{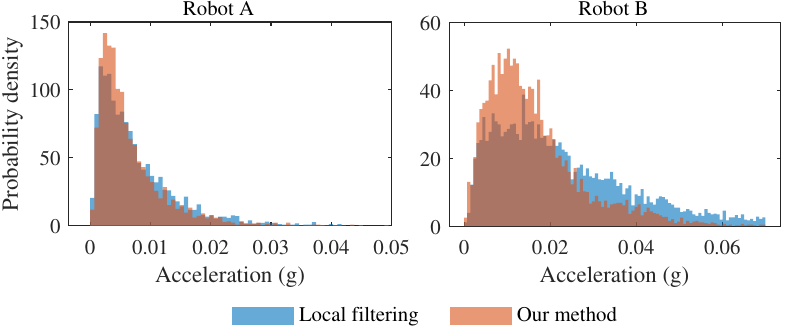}
    \caption{The acceleration distribution of Robot A's end-effector and Robot B's end-plate during manufacturing in Example II with $g$ representing the magnitude of the gravitational acceleration.}
    \label{fig:Ex2AccSensor}
\end{figure}

\begin{figure}[!t]
    \centering
    \includegraphics[width=.5\textwidth]{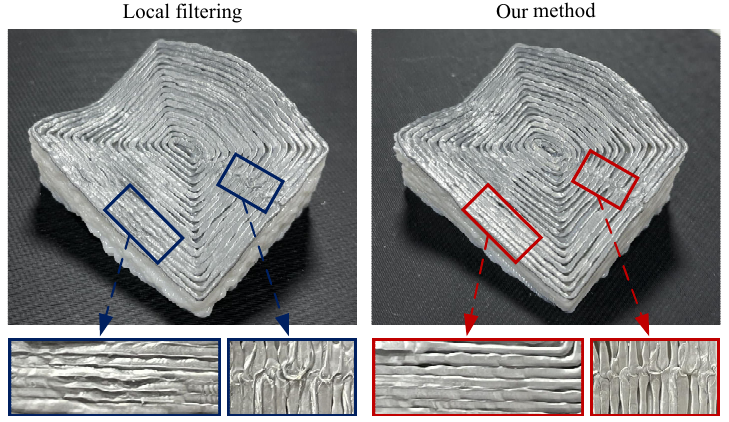}
    \caption{The manufacturing results of Example II using trajectories optimized by different methods.} 
    \label{fig:Ex2Exp}
\end{figure}

Similar to the earlier example, we also conducted physical fabrication to verify the effectiveness of our trajectory optimization method. Vibrations at the end-effectors of robots A and B are measured by accelerometers, and the results are illustrated in Table \ref{Table:VibrationEx2} and Fig.~\ref{fig:Ex2AccSensor}. By using our method, the maximum and average acceleration of both Robot A end and Robot B end are decreased compared with the local filtering method. The improvement of the Robot B is relatively more pronounced, the average acceleration of which decreases by 30.96\%. This is consistent with the smoothness performance of joint paths shown in Fig.~\ref{fig:Ex2jointJerk}. The 3D printed curve layers are as shown in Fig. \ref{fig:Ex2Exp}. It can be found that our method can further enhance surface quality due to the improvement of joint jerks.

\subsection{Example III}
\begin{figure}[!t]
\centering
\includegraphics[width=.41\textwidth]{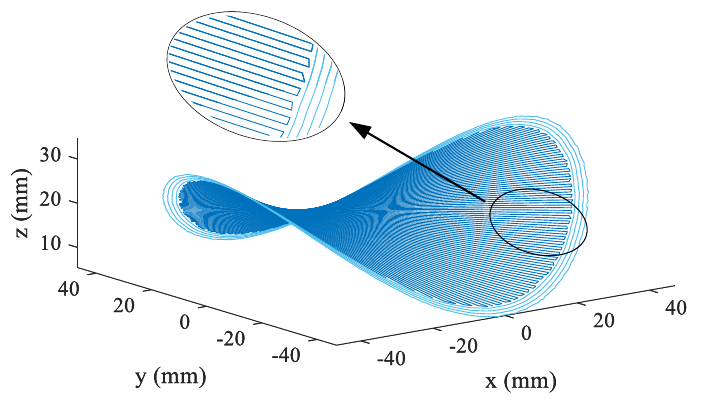}
\vspace{-8pt}
\caption{The toolpath of a curved layer for AM employed in Example III.}
\label{fig:Ex3Path}
\end{figure}

The third example is to use our method for optimizing the geometric smoothness of the trajectory\footnote{Note that this only changes the joint path but not the 3D toolpath.}. As discussed in Sec.~\ref{Sec:DiscuGenera}, this can be achieved by replacing $\{t_i \}$ in the optimization problem with a set of fixed $\{s_i \}$ and excluding time information from the optimization process. 
The curved layer to fabricate is a saddle surface with a zig-zag toolpath containing $8,832$ waypoints as shown in Fig.~\ref{fig:Ex3Path}. The optimization result is listed in Table \ref{Table:solution3}, showing that all three geometric smoothness metrics have been enhanced after optimization.

\begin{table}[!t]
\begin{center}
\caption{Optimization results in Example III.}
\label{Table:solution3}
\begin{tabular}{|c|c|c|r|r|}
\hline
\multicolumn{3}{|c|}{} & Before Optm. &  After Optimization \\
\hline
\hline
\multirow{4}{*}{$|\frac{dq}{ds}|$ } & Maximal & Robot A      & 18.90 & 4.06 \\
                            & value   & Robot B & 1518.09 & 145.54 \\ \cline{2-5}
                            & Average & Robot A     & 1.50 & 1.00\\
                            & value   & Robot B     & 16.15 & 13.55\\
\hline
\multirow{4}{*}{$|\frac{d^2q}{ds^2}|$ } & Maximal & Robot A     & 519.63 & 6.44 \\
                            & value   & Robot B& 3652.33 & 167.45 \\ \cline{2-5}
                            & Average & Robot A     & 0.17 & 0.11\\
                            & value   & Robot B & 3.00 & 1.73\\
\hline
\multirow{4}{*}{$|\frac{d^3 q}{ds^3}|$ } & Maximal & Robot A    & 66.23 & 6.68 \\
                            & value   & Robot B & 759.07 & 130.96 \\ \cline{2-5}
                            & Average & Robot A     & 0.12 & 0.07\\
                            & value   & Robot B & 2.32 & 1.12\\
\hline
\end{tabular}
\begin{flushleft}
\item$^*$Here $|{dq}/{ds}|$, $|{d^2q}/{ds^2}|$ and $|{d^3 q}/{ds^3}|$ represent the first-, second-, and third-order derivatives of the joint angle w.r.t. the path arc-length parameter.
\end{flushleft} 
\end{center}
\end{table}

\begin{figure}[!t]
\centering
\includegraphics[width=.45\textwidth]{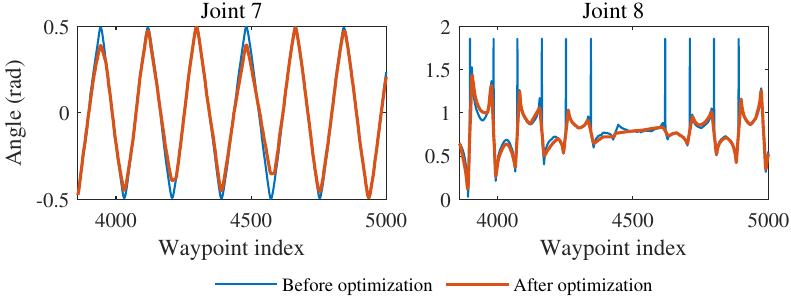}
\vspace{-5pt}
\caption{The joint paths of Robot B near the singularity in Example III.}
\label{fig:Ex3Joints}
\end{figure}

One benefit of our method is that it can effectively avoid the drastic shaking of Joint 8 (i.e., the second joint of the position table as Robot B) caused by singularity. This has been demonstrated in Fig. \ref{fig:Ex3Joints}, which shows the comparison of the joint paths of Robot B near its singular region (i.e., where the first joint angle of Robot B is near zero). 

\begin{figure}[!t]
\centering
\includegraphics[width=.4\textwidth]{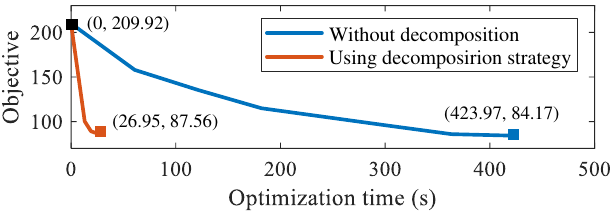}
\vspace{-5pt}
\caption{Effectiveness study of the decomposition scheme by Example III.}
\label{fig:Ex3Decom}
\end{figure}

The process optimization with and without the decomposition scheme has been tested on the first 500 waypoints with the comparison shown in Fig. \ref{fig:Ex3Decom}. By using the decomposition scheme, the computational efficiency can be improved by 93.64\% while obtaining the result with similar quality.

\begin{figure}[!t]
\centering
\includegraphics[width=\linewidth]{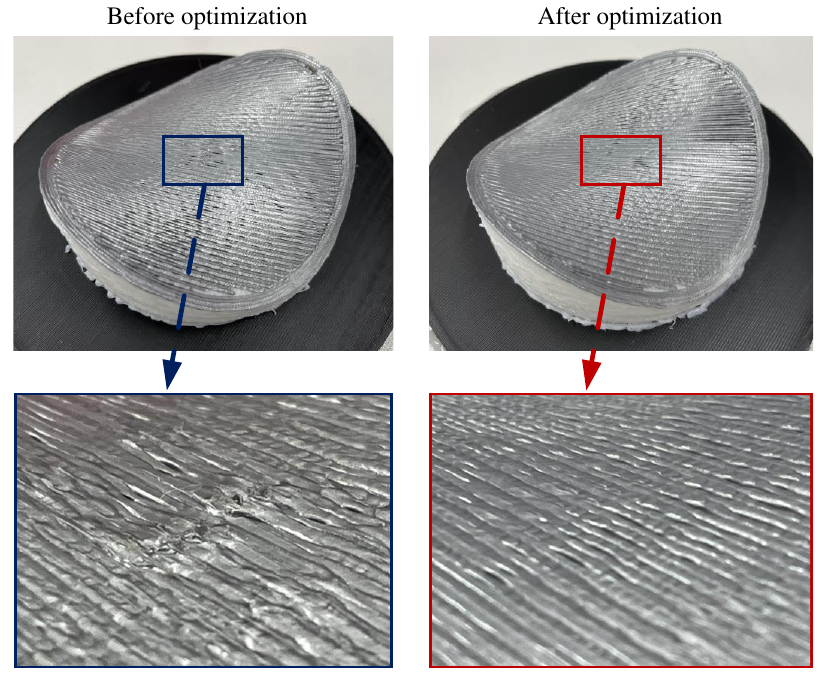}
\caption{The manufacturing results of Example III using trajectories before vs. after optimization.}
    \label{fig:Ex3Exp}
\end{figure}

The functionality of the improved smoothness on joint paths can be demonstrated by the quality of physical fabrication. Both unoptimized and optimized trajectories are fed into the controller of the robots, and the maximum allowable tooltip speed is set to $20mm/s$. The real tooltip speed is planned by the robot controller automatically according to the kinematic and dynamic limits. The manufacturing time recorded shows that the printing time is reduced from 587~sec. to 534~sec. by optimization. The 3D printed parts are as shown in Fig. \ref{fig:Ex3Exp}, where noticeable improvement of surface quality can be observed -- especially when Robot B is near the singularity zone. 

\begin{figure}[!t]
\centering
\includegraphics[width=.3\textwidth]{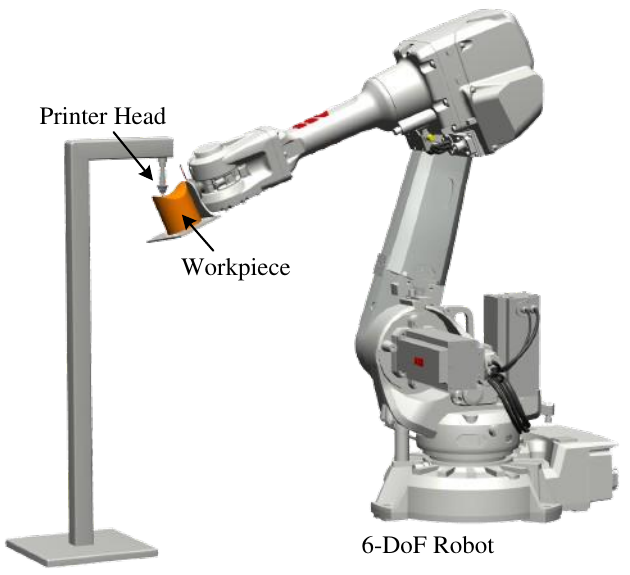}
\caption{A single robot system with 6-DoF for AM as Example IV.}
\label{fig:newRob}
\end{figure}
\subsection{Example IV}
In this example, we optimize the trajectory to realize the surface and toolpath of Example III on a single robot system with 6-DoF as shown in Fig.~\ref{fig:newRob}, where a fixed printer head is employed with the similar configuration as \cite{Dai2018}. 

\begin{figure}[!t]
\centering
\includegraphics[width=.44\textwidth]{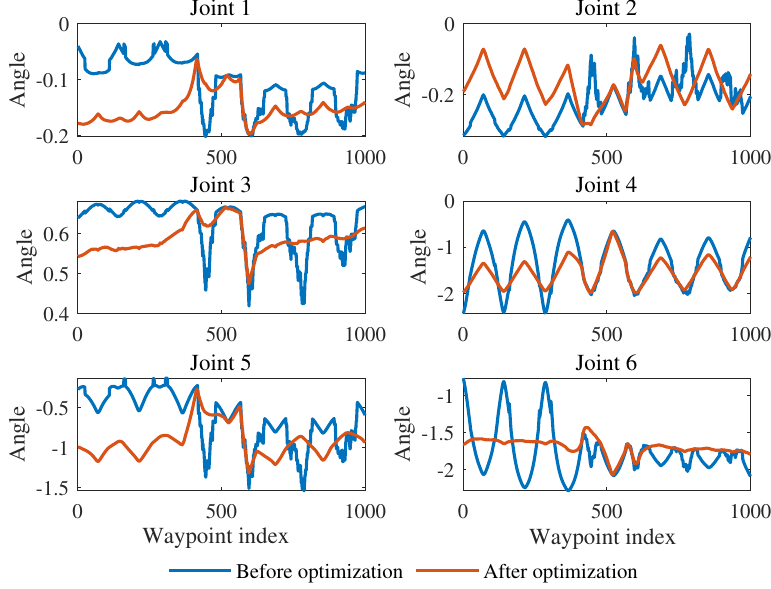}
\caption{The robot joint paths of Example IV.}
\label{fig:Ex4Joints}
\end{figure}

\begin{figure}[!t]
    \centering
    \includegraphics[width=.44\textwidth]{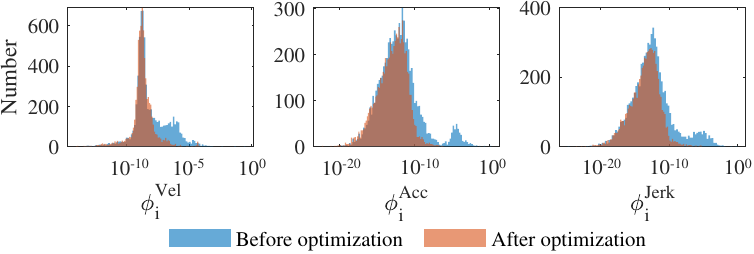}
    \caption{Histograms of smoothness metrics for Example IV.}
    \label{fig:Ex4Phi}
\end{figure}

The comparison of the joint paths using the first 1000 waypoints is shown in Fig. \ref{fig:Ex4Joints}, indicating that optimization improves the trajectory smoothness. The performance improvement is also verified by the histograms of kinematic smoothness as given in Fig.~\ref{fig:Ex4Phi}. The effectiveness has also been demonstrated by the comparison shown in the supplementary video.
\section{Conclusion and Discussion}
\label{sec:conclusion}
This paper presents a novel concurrent trajectory optimization framework for robot-assisted manufacturing. Based on the kinematic model of the dual robot system, trajectory optimization models are constructed considering the special requirements for the manufacturing process. Our computational framework can concurrently optimize the tool orientation, the kinematic redundancy, and the manufacturing time-sequence while minimizing the kinematic smoothness metrics. With the help of a newly proposed decomposition-based numerical scheme, the quality of trajectories with a large number of waypoints can be effectively improved with high efficiency. 

The performance of our framework has been demonstrated on different toolpaths to fabricate freeform surfaces. Both simulations and physical experiments are conducted for the verification. Compared with the unoptimized trajectories and the results of the local filtering method \cite{Dai2020}, our method can achieve much better kinematic smoothness, resulting in higher surface quality of physical fabrication. Meanwhile, the computing time spent on optimization can be reduced by more than 60\% compared with the local filtering method and over 90\% compared with the computation without applying the decomposition scheme.

In our current implementation, the final collision correction step may affect the optimality of the results although it rarely occurs during our experimental tests. \rev{}{Moreover, it takes a relatively long time with up to 5 minutes to learn a high-quality proxy function for collision detection, while for large-scale paths with up to 8k waypoints, the optimization process itself takes less than 10 minutes. This is considered as a major limitation of our approach as the proxy function needs to be re-learned when the environment or the model changes.} A \rev{}{more adaptive} collision detection technique is needed for future research. 

\section*{Acknowledgments}
The project is partially supported by the UK Engineering and Physical Sciences Research Council (EPSRC) Fellowship Grant (Ref.\#: EP/X032213/1) and the chair professorship fund at the University of Manchester.


{\appendices

\section{Formulas for joint acceleration and jerk} \label{AppA}
The acceleration of joint obtained by the unevenly spaced numerical differentiation is
\begin{equation}
    \textbf{a}_i = \frac{2 \textbf{q}_{i-1}}{t_i (t_i + t_{i+1} )} - \frac{2 \textbf{q}_{i}}{t_i t_{i+1}} + \frac{2 \textbf{q}_{i+1}}{t_{i+1} (t_i + t_{i+1} )}
\end{equation}
The joint jerk can be evaluated by
\begin{equation}
    \textbf{j}_i = \rho^1_i \textbf{q}_{i-2} + \rho^2_i \textbf{q}_{i-1} +\rho^3_i \textbf{q}_{i}+\rho^4_i \textbf{q}_{i+1}+\rho^5_i \textbf{q}_{i+2}
\end{equation}
with
\begin{equation}
    \rho^1_i = \frac {-12t_{i+1}-6t_{i+2}+6t_i} {t_{i-1} (t_{i-1} + t_i ) (t_{i-1} + t_i + t_{i+1}) (\sum_{j=i-1}^{i+2} t_j) },
\end{equation}
\begin{equation}
    \rho^2_i = \frac {12t_{i+1}+6t_{i+2}-6t_i-6t_{i-1}} {t_{i-1} t_i (t_i + t_{i+1}) (t_i + t_{i+1} + t_{i+2}) },
\end{equation}
\begin{equation}
    \rho^3_i = \frac {-12t_{i+1}-6t_{i+2}+12t_i+6t_{i-1}} {t_i t_{i+1} (t_{i+1} + t_{i+2}) (t_{i-1} + t_i) },
\end{equation}
\begin{equation}
    \rho^4_i = \frac {6t_{i+1}+6t_{i+2}-12t_i-6t_{i-1}} {t_{i+1} t_{i+2} (t_i + t_{i+1}) (t_{i-1} + t_{i} + t_{i+1}) },
\end{equation}
\begin{equation}
    \rho^5_i = \frac {-6t_{i+1}+12t_i + 6t_{i-1}} {t_{i+2} (t_{i+1} + t_{i+2} ) (t_{i} + t_{i+1} + t_{i+2}) (\sum_{j=i-1}^{i+2} t_j) }.
\end{equation}
The formulas are derived according to \cite{Gautschi2011}.

\section{Formulas of tooltip acceleration}\label{AppB}
Denote the velocity of the tooltip as $\textbf{v}^{TTP}=v\textbf{t}$, where $\textbf{t}$ is the unit tangent of the toolpath. The acceleration at the tooltip can be computed by
\begin{equation}
    \begin{split}
        \textbf{a}^{TTP} &= \frac{d \textbf{v}^{TTP}}{d t} = \frac{dv}{dt} \textbf{t} + v \frac{d\textbf{t}}{dt}\\
        &= \frac{dv}{dt} \textbf{t} + \kappa v^2 \textbf{n}
    \end{split}
\end{equation}
where $\kappa$ and $\textbf{n}$ represent the curvature and the unit normal of the toolpath. The acceleration consists of the tangential part and the normal part. 

For the toolpath in a discrete form, the tangential acceleration $a_i^t=dv/dt$ at the $i$th waypoint is
\begin{equation}
    \begin{split}
        a_i^t &=\frac{dv}{dt} =\frac{v_{i+1}^t-v_{i}^t}{(t_i+t_{i+1})/2}\\
        &= \frac{2(s_{i+1}t_i-s_it_{i+1})}{t_it_{i+1}(t_i+t_{i+1})}
    \end{split}   
\end{equation}
with $v_i^t=s_i/t_i$. As illustrated in Fig.\ref{fig:TTPAcc}, let
\begin{equation}
    \Bar{\textbf{v}}_i^t=\frac{(\textbf{p}_i - \textbf{p}_{i-1})/t_i+(\textbf{p}_{i+1} - \textbf{p}_{i})/t_{i+1}}{2},
\end{equation}
the discrete form of the normal acceleration $a_i^n= \kappa v^2$ is obtained as 
\begin{equation}
    \begin{split}
        a_i^n &= \kappa_i \left\| \Bar{\textbf{v}}_i^t \right\| ^2 \\
        &= \kappa_i \frac
    {s_i^2 t_{i+1}^2 +s_{i+1}^2 t_i^2 -2 s_i s_{i+1} t_i t_{i+1} \textup{cos}(\sigma_i) } 
    {4 t_i^2 t_{i+1}^2}
    \end{split}    
\end{equation}
where $\kappa_i$ and $\sigma_i$ are the path curvature and the angle at $\mathbf{p}_i$.

\begin{figure}
    \centering
    \includegraphics[width=0.15\textwidth]{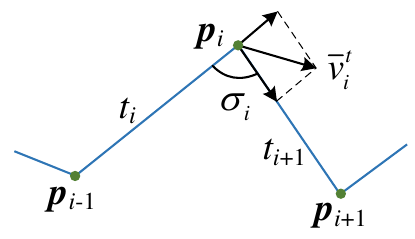}
    \caption{The local schematic of the toolpath in a discrete form.}
    \label{fig:TTPAcc}
\end{figure}

\bibliographystyle{IEEEtran}
\bibliography{reference}{}

\begin{IEEEbiography}[{\includegraphics[width=1in,height=1.25in,clip,keepaspectratio]{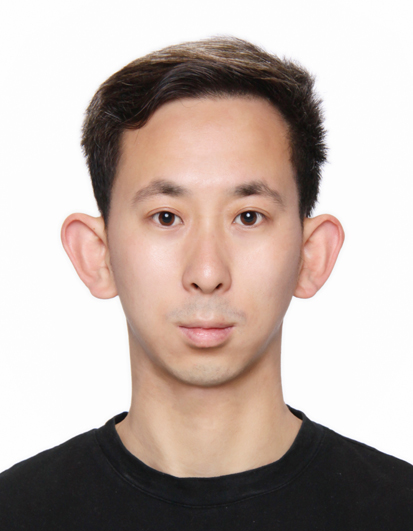}}]{Yongxue Chen}
received the B.Eng. and M.Eng. degrees in mechanical engineering from Shanghai Jiao Tong University, Shanghai, China, in 2020 and 2023, respectively. He is currently working towards the Ph.D. degree in robot-assisted manufacturing with the Digital Manufacturing Lab, School of Engineering, The University of Manchester, Manchester, U.K.

His current research interests include robotics and additive manufacturing.
\end{IEEEbiography}

\begin{IEEEbiography}[{\includegraphics[width=1in,height=1.25in,clip,keepaspectratio]{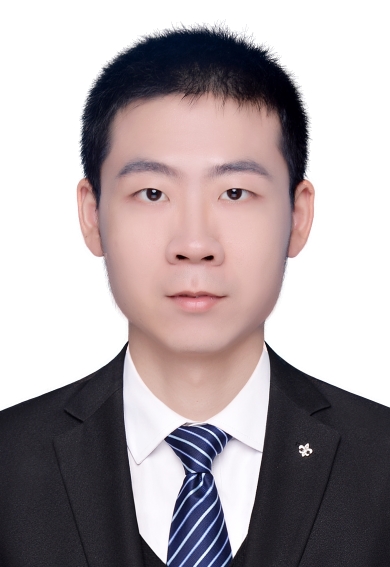}}]{Tianyu Zhang}
(Student Member, IEEE) obtained his B.Eng. degree in mechanical engineering from the University of Electronic Science and Technology of China, China, in 2015, followed by a Master's degree from Xi’an Jiaotong University, China, in 2018. He recently completed his Ph.D. in mechanical engineering at the University of Manchester, UK, in 2024 and is currently engaged in further research within the same group.

His research focuses on robotic-assisted multi-axis additive manufacturing, geometry computing, and composite fabrication.
\end{IEEEbiography}

\begin{IEEEbiography}[{\includegraphics[width=1in,height=1.25in,clip,keepaspectratio]{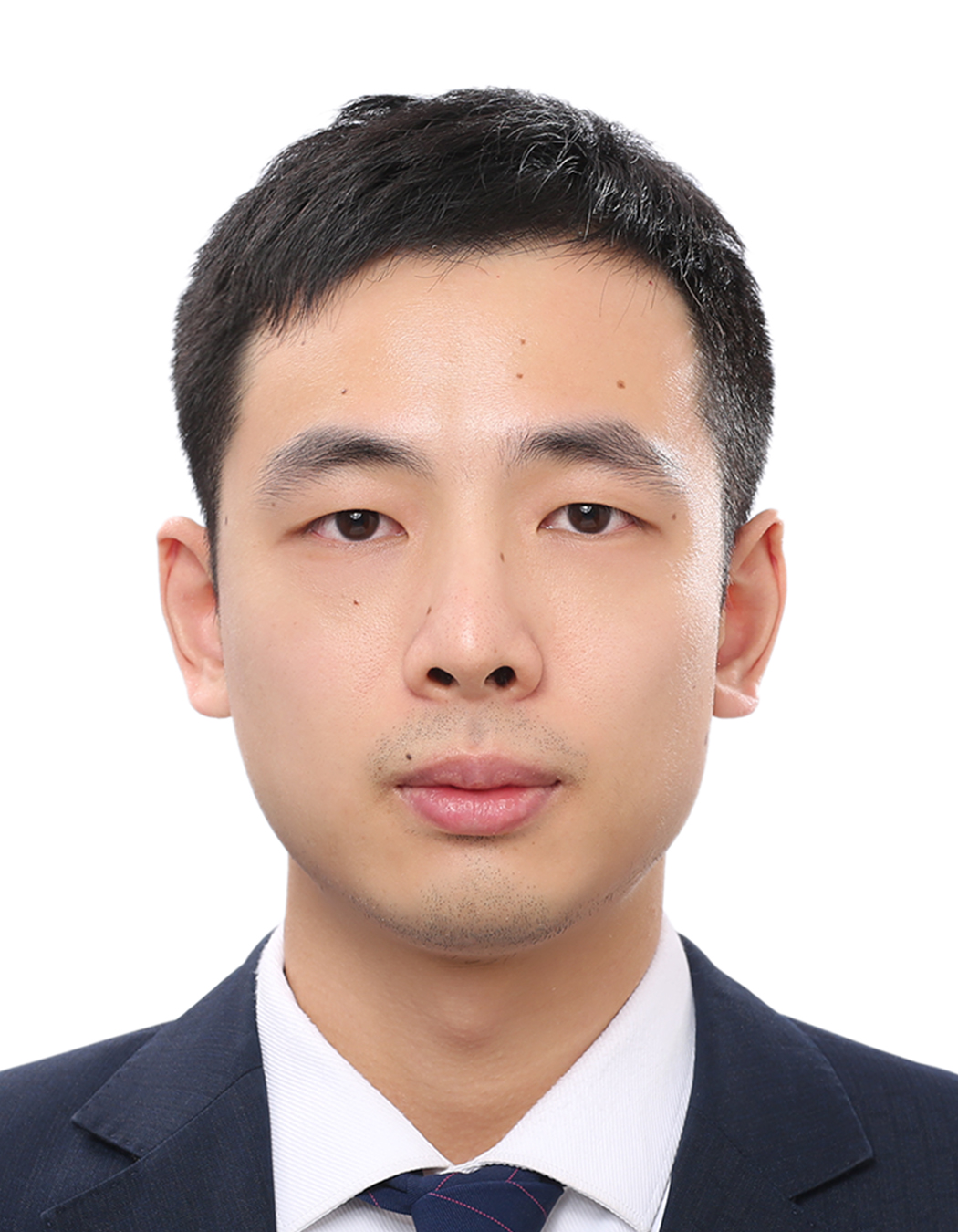}}]{Yuming Huang}
received the B.Eng. degree in process equipment and control engineering from the Taiyuan University of Technology, Taiyuan, China, in 2018 and the Msc degree in mechanical and automation engineering from the Chinese University of Hong Kong, Hong Kong, China, in 2019. From 2020 to 2021, he worked as a mechanical engineer at Changzhou Architectural Research Institute Group Co., Ltd., Changzhou, China. He is currently working toward the Ph.D. degree at the Digital Manufacturing Lab, The University of Manchester, Manchester, U.K. 

His research interest includes computation design for additive manufacturing, computation fabrication, and robot-assisted 3D printing.
\end{IEEEbiography}

\begin{IEEEbiography}[{\includegraphics[width=1in,height=1.25in,clip,keepaspectratio]{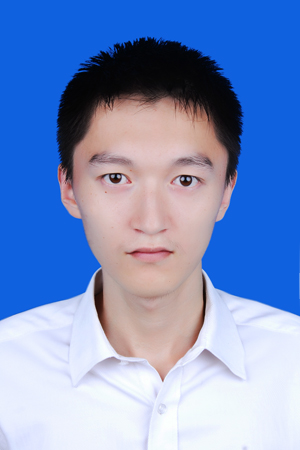}}]{Tao Liu}
is a Ph.D. student at the University of Manchester in the School of Engineering. He received his master's degree in the School of Mathematics and Statistics at Central South University, China, in 2020.

His research interests include implicit surface modeling and robot assistant 3D printing.
\end{IEEEbiography}

\begin{IEEEbiography}[{\includegraphics[width=1in,height=1.25in,clip,keepaspectratio]{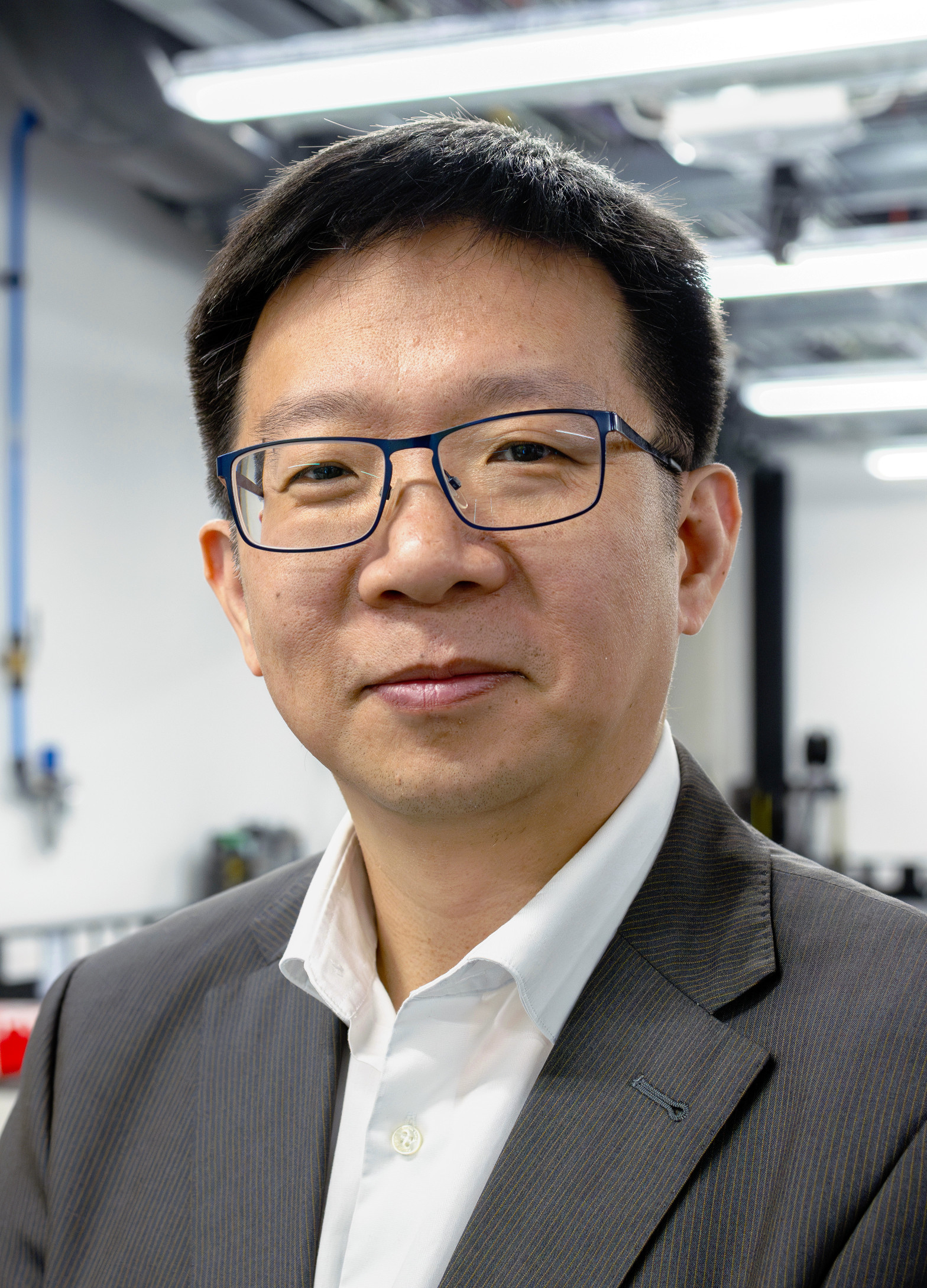}}]{Charlie C.L. Wang}
(Senior Member, IEEE) received the B.Eng. degree in mechatronics engineering from the Huazhong University of Science and Technology, China, in 1998 and the Ph.D. degree in mechanical engineering
from the Hong Kong University of Science and Technology, Hong Kong, in 2002. He is currently a Professor and Chair of Smart Manufacturing at the University of Manchester, United Kingdom. He was elected as a Fellow of the American Society of Mechanical Engineers in 2013.

His research interests include digital manufacturing, computational design, soft robotics, mass personalization, and geometric computing. 
\end{IEEEbiography}

\vfill

\end{document}